\documentclass[conference,compsoc]{IEEEtran}

\ifCLASSOPTIONcompsoc
  \usepackage[nocompress]{cite}
\else
  \usepackage{cite}
\fi
\usepackage{amsmath, amssymb}   
\usepackage{booktabs}           
\usepackage{graphicx}           
\usepackage{multirow}           
\usepackage{microtype}          
\usepackage{caption} 
\usepackage{pifont}
\usepackage{enumitem}

\usepackage{hyperref}       
\usepackage{url}
\usepackage{amsthm}
\usepackage{float}

\usepackage{arydshln}
\usepackage{xcolor}
\usepackage{booktabs}

\ifCLASSINFOpdf
\else
\fi
\ifCLASSOPTIONcompsoc
 \usepackage[caption=false,font=footnotesize,labelfont=sf,textfont=sf]{subfig}
\else
 \usepackage[caption=false,font=footnotesize]{subfig}
\fi

\newcommand{\bheading}[1]{\vspace{2pt}\noindent\textbf{#1}}
\begin{document}

\title{A Systematic Study of Model Extraction Attacks on Graph Foundation Models}

\IEEEoverridecommandlockouts
\author{
\IEEEauthorblockN{
Haoyan Xu\textsuperscript{$\spadesuit$}\IEEEauthorrefmark{1},
Ruizhi Qian\textsuperscript{$\spadesuit$}\IEEEauthorrefmark{1},
Jiate Li\IEEEauthorrefmark{1},
Yushun Dong\IEEEauthorrefmark{2},
Minghao Lin\IEEEauthorrefmark{1},
Hanson Yan\IEEEauthorrefmark{1},\\
Zhengtao Yao\IEEEauthorrefmark{1},
Qinghua Liu\IEEEauthorrefmark{3},
Junhao Dong\IEEEauthorrefmark{4},
Ruopeng Huang\IEEEauthorrefmark{1},
Yue Zhao\textsuperscript{$\heartsuit$}\IEEEauthorrefmark{1},
Mengyuan Li\textsuperscript{$\heartsuit$}\IEEEauthorrefmark{1}
}

\IEEEauthorblockA{\IEEEauthorrefmark{1}University of Southern California \quad
\IEEEauthorrefmark{2}Florida State University}
\IEEEauthorblockA{\IEEEauthorrefmark{3}The Ohio State University \quad
\IEEEauthorrefmark{4}Nanyang Technological University}

\thanks{\textsuperscript{$\spadesuit$}Equal contribution.}
\thanks{\textsuperscript{$\heartsuit$}Corresponding authors.}
}

\maketitle

\begin{abstract}
 
Graph machine learning (GML) has advanced rapidly in tasks such as link prediction, anomaly detection, and node classification. As models scale up, pretrained graph models have become valuable intellectual assets, embodying extensive computation and domain expertise.
Building on these advances, the emergence of Graph Foundation Models (GFMs) marks a major leap by jointly pretraining graph and text encoders on massive, heterogeneous data, unifying structural and semantic understanding within one framework. This enables zero-shot inference and cross-domain generalization, powering applications such as fraud detection and biomedical analysis. However, at the same time, the vast pretraining cost and cross-domain knowledge embedded in GFMs make them highly valuable intellectual property, thereby turning well-trained GFMs into increasingly attractive targets for model extraction attacks (MEAs). Yet, prior MEA research has focused only on small-scale graph neural networks (GNNs) trained on a single graph, leaving the security implications for large-scale and multimodal GFMs largely unexplored.

This paper presents the first systematic investigation of MEAs against GFMs. We begin by formalizing a realistic black-box threat model and develop a comprehensive taxonomy of six practical attack scenarios, spanning domain-level and graph-specific extraction goals, architectural mismatch, limited query budgets, partial node access, and training data discrepancies.
To instantiate these attacks, we introduce a lightweight yet effective extraction framework based on supervised regression of graph embeddings. Remarkably, even without access to contrastive pretraining data, this method learns an attacker encoder that aligns closely with the victim’s text encoder and retains its zero-shot inference capability on unseen graphs.
Theoretical analysis and experiments on seven datasets show that the attacker can approximate the victim model using only 0.07\% of its original training time, with an average classification accuracy gap of just 0.0015.
Overall, our findings reveal that GFMs substantially expand the model extraction attack surface, underscoring the urgent need for deployment-aware security defenses in large-scale multimodal graph learning systems.
\end{abstract}


%
\maketitle
\thispagestyle{plain}
\pagestyle{plain}
\IEEEpeerreviewmaketitle

\section{Introduction}
In recent years, graph machine learning (GML) has achieved strong performance across a range of tasks~\cite{duan2022multivariate, xu2020multivariate}, including link prediction~\cite{xiao2020timme}, anomaly detection~\cite{liu2022bond, xu2025few, xu2025graph}, and node classification~\cite{xu2024lego}. This success has driven the adoption of GML models in domains ranging from social networks and recommender systems to biological networks and financial transaction graphs~\cite{xiao2020timme, xu2020cosimgnn, xu2021graph}. However, training modern GML models requires substantial computational resources as both model complexity and graph size continue to grow. As a result, pretrained GML models have become valuable intellectual property~\cite{li2025intellectual} and attractive targets for adversaries. A prominent threat in this context is model extraction attacks (MEAs), which aim to recover a model’s functionality by training a surrogate to mimic the target’s outputs. Prior work shows that with enough queries, an attacker can effectively duplicate a target model’s knowledge into a compact substitute. Such extraction of proprietary graph models raises serious privacy and intellectual-property concerns, since an attacker can reproduce specialized services without authorization.

\bheading{Model Extraction on Graph Neural Networks.} Despite the rising stakes, existing research on graph model extraction has been limited to relatively small-scale scenarios. Several studies have examined MEAs against graph neural networks (GNNs) for node-level tasks~\cite{wu2022model, defazio2019adversarial, zhuang2024unveiling, shen2022model, guan2024realistic}, but these efforts focused on modest GNN architectures (e.g., GCN~\cite{kipf2016semi}, GAT~\cite{velivckovic2017graph}, GraphSAGE~\cite{hamilton2017inductive}) trained on a single graph. These works explored various practical constraints, including attackers with access to only partial graph structure or node features~\cite{wu2022model}, inductive (unseen-node) settings~\cite{shen2022model}, query and batch-size limitations~\cite{wang2025cega}, and even data-free extraction~\cite{zhuang2024unveiling}. In such single-graph settings, the surrogate is optimized solely for that particular graph and cannot generalize beyond it. This leaves a critical open question: \textit{Are the new generation of large-scale, cross-domain Graph Foundation Models equally vulnerable to extraction?}

\bheading{Threats to Graph Foundation Models.} In this work, we answer the above question by conducting the first systematic study of model extraction attacks on Graph Foundation Models (GFMs). GFMs are a new class of large-scale, multimodal graph models that differ fundamentally from traditional GNNs in both architecture and capability~\cite{zhu2025graphclip, chen2024text, liu2023one}. In particular, a GFM is jointly pretrained on massive heterogeneous data by aligning a graph encoder with a text encoder, fusing structural graph information with semantic textual knowledge. This yields powerful cross-domain generalization and zero-shot inference abilities that single-graph GNNs lack. As a result, GFMs have emerged as critical infrastructure in high-value applications such as financial fraud detection, biomedical knowledge graph analysis, cybersecurity threat intelligence, and large-scale recommendation systems. Paradoxically, however, the very generality that makes GFMs so powerful also creates a much broader and more vulnerable attack surface than traditional GNNs. Unlike a conventional graph-specific GNN restricted to a single graph, a GFM provides general-purpose graph-text understanding across domains. Thus, successfully extracting a GFM compromises not only its task performance and graph embeddings, but also its cross-domain knowledge base and zero-shot inference ability. In practice, an attacker who steals a GFM obtains a single surrogate model capable of performing a wide array of tasks on new graphs (e.g., unauthorized link prediction or node classification) without any additional training—far exceeding the impact of stealing a GNN trained on a single graph. These unique properties introduce new vulnerabilities and demand new extraction techniques, motivating us to rethink the attack surface beyond prior studies on small graph-specific GNNs.

\bheading{Unified Extraction Methodology and Taxonomy for GFMs.} In this paper, we present the first systematic investigation of MEAs on GFMs, centered on state-of-the-art representative multimodal GFM as the target victim model. At the core of our approach is a novel and efficient attack technique tailored specifically to GFMs. Rather than replicating the GFM’s complex multimodal pretraining regimen, we train a surrogate graph encoder to regress the victim GFM’s output embeddings on some accessible graph data using a simple supervised loss. This one-loss embedding-regression strategy allows the adversary to steal the GFM’s functionality with minimal cost. Notably, our surrogate closely aligns with the victim’s latent representations without requiring any proprietary graph-text training pairs or knowledge of the GFM’s internal architecture. Consequently, even without using the original contrastive graph-text training data, the stolen surrogate model retains the victim’s alignment with the text encoder and thus preserves its zero-shot inference capabilities on unseen graphs. In essence, the attacker reconstructs the core behavior of the target GFM through a significantly cheaper and more data-efficient procedure.

To rigorously assess GFM vulnerabilities, we also introduce a principled evaluation framework for model extraction. We assume a realistic black-box threat model in which the adversary can query the victim’s graph encoder and observe its output embeddings, under various practical constraints (e.g., limited queries, partial knowledge of the graph’s nodes or features, no access to original contrastive training corpora). Within this threat model, we define a taxonomy of six representative attack scenarios capturing different adversarial assumptions. These scenarios vary along four key axes: (i) the adversary’s extraction goal (full cross-domain vs.\ single-graph), (ii) the surrogate’s capacity and architecture (architecture matching the victim vs.\ mismatched or lightweight), (iii) the query budget (unrestricted vs.\ strictly limited), and (iv) the attacker’s available data (complete vs.\ partial or mismatched graph data). This taxonomy provides a structured framework for exploring how various real-world conditions impact the success of extracting a GFM, ensuring that our analysis covers the most relevant threat scenarios.

\bheading{Contributions.}
We implement and evaluate our attack framework across six attack scenarios. Our results show that under diverse, realistic conditions—and using only a single embedding-regression loss—an adversary can train high-fidelity surrogates that preserve the victim’s zero-shot performance on unseen graphs. These findings indicate that modern GFMs face considerable extraction risk even when the adversary lacks the original contrastive pretraining data and even when the surrogate model has a different architecture or far fewer parameters. Remarkably, we successfully steal the full victim GFM using only $\sim$0.07\% of its original training time, with the surrogate’s average accuracy across seven evaluation datasets reduced by only 0.0015. Our main contributions are summarized as follows:

\begin{itemize}[leftmargin=*,itemindent=0pt]
\item \textbf{Scope.} We present the first comprehensive analysis of model extraction attacks on GFMs, moving beyond prior work that was limited to small-scale, graph-specific GNNs.

\item \textbf{Threat model \& taxonomy.} We formalize a realistic black-box threat model for multimodal GFMs and introduce a six-scenario taxonomy covering variations in attacker goals (full cross-domain vs.\ single-graph), surrogate model capacity/architecture (matched vs.\ mismatched), query budget (unlimited vs.\ constrained), and data visibility (complete vs.\ partial/mismatched).

\item \textbf{One-loss embedding attack.} We propose a novel one-loss embedding-regression attack (MEA-GFM) that trains a surrogate using a single MSE loss to match the victim’s graph encoder outputs. We provide a theoretical explanation showing that aligning these embeddings is sufficient to preserve the victim’s zero-shot inference capabilities via the shared text encoder, which explains why contrastive pretraining data is unnecessary for our attack.

\item \textbf{Practical risk.} We demonstrate the practical risk to GFMs: across all six scenarios and seven datasets, the extracted surrogates achieve high fidelity and near-victim accuracy while requiring only a tiny fraction of the original training cost. 
\end{itemize}

\begin{figure*}[!t]
   \begin{center}
      \includegraphics[width=0.95\linewidth]{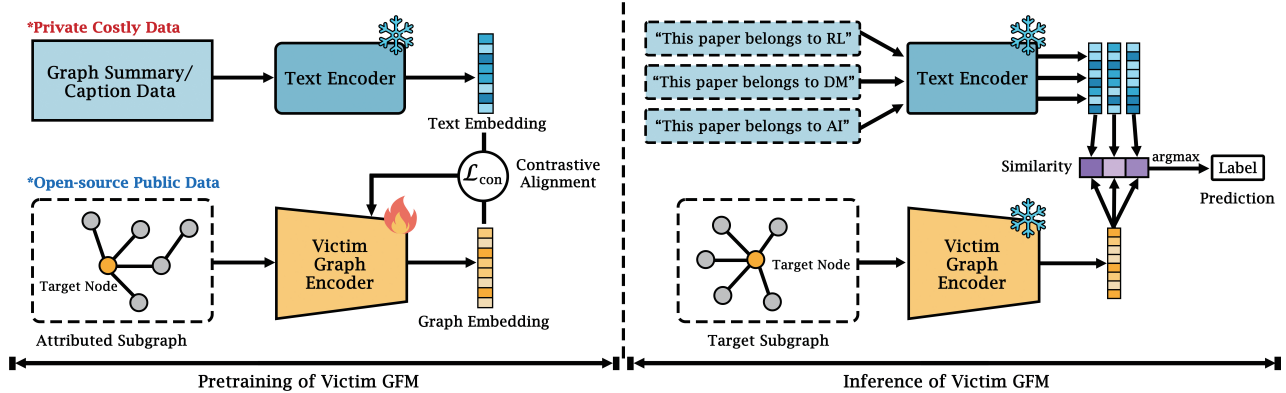} %
    \end{center}
 \caption{Overview of the GFM. (1) The victim GFM is pretrained using contrastive learning on attributed subgraphs paired with graph-summary or caption data. Only the graph encoder is trainable; the text encoder remains frozen. (2) During inference, the pretrained graph encoder maps a target subgraph into a shared embedding space with the text encoder. Zero-shot node classification is performed by computing similarity between the graph embedding and textual label embeddings, and selecting the label with the highest similarity score.}
\label{fig:GFM}
\end{figure*}
\section{Background}
\label{sec:preliminary}
\subsection{Text-Attributed Graph}
A text-attributed graph (TAG) \cite{yan2023comprehensive} is defined as
\( G = (\mathcal{V}, \mathbf{A}, \mathcal{M}, \mathbf{X}) \),
where \( \mathcal{V} = \{v_1, \dots, v_n\} \) is the node set, and each node is associated with a raw text attribute (e.g., a document, product description, or profile field).
The collection of textual attributes is represented as \( \mathcal{M} = \{m_1, m_2, \dots, m_n\} \),
which are transformed into dense vector embeddings \( \mathbf{X} = \{x_1, x_2, \dots, x_n\} \) using a pretrained language model \cite{wang2020minilm, reimers2019sentence}.
The adjacency matrix \( \mathbf{A} \in \{0,1\}^{n \times n} \) encodes the graph’s structure, where
\( \mathbf{A}[i, j] = 1 \) indicates the presence of an edge between nodes \( v_i \) and \( v_j \).

\subsection{Graph Neural Networks}
GNNs are a class of deep learning models designed to operate on graph-structured data.
Given a graph \( G \) with node features \( \mathbf{X} \) and adjacency matrix \( \mathbf{A} \), GNNs learn node representations by recursively aggregating information from neighboring nodes and edges.
A typical message-passing layer can be expressed as:
\begin{equation}
    \mathbf{H}^{(l+1)} = \sigma \!\left( \tilde{\mathbf{A}}\, \mathbf{H}^{(l)} \mathbf{W}^{(l)} \right),
\end{equation}
where \( \tilde{\mathbf{A}} \) is the normalized adjacency matrix with self-loops, \( \mathbf{W}^{(l)} \) is the learnable weight matrix at layer \( l \), \( \mathbf{H}^{(l)} \) denotes node representations, and \( \sigma(\cdot) \) is a nonlinear activation function.
After multiple layers, node embeddings capture higher-order structural and feature information, which can then be aggregated for downstream tasks such as node classification, link prediction, or graph-level prediction.
Representative architectures include GCN~\cite{kipf2016semi}, GraphSAGE~\cite{hamilton2017inductive}, and GAT~\cite{velivckovic2017graph}, which differ in their neighborhood aggregation and attention mechanisms.

\subsection{Graph Foundation Model Overview}
In this work, we investigate the security vulnerabilities of large-scale graph models by conducting MEAs on a representative text-graph foundation model, GraphCLIP \cite{zhu2025graphclip}.
Similar to other multimodal foundation models such as CLIP \cite{radford2021learning}, AudioCLIP \cite{guzhov2022audioclip}, and PointCLIP \cite{zhang2022pointclip}, GraphCLIP is pretrained on a diverse set of datasets from multiple domains using large-scale contrastive learning, as illustrated in Fig.~\ref{fig:GFM}. This training aligns the attributed graph space with the textual semantic space.
It consists of a text encoder and a graph encoder that jointly learn cross-modal representations.
\textbf{During pretraining, only the graph encoder is trainable, while the text encoder remains frozen.}
After pretraining, GraphCLIP enables zero-shot inference on unseen datasets for tasks such as node classification and link prediction.
Below, we briefly introduce the key components of GraphCLIP.

\noindent \textbf{Graph Encoder.}
For each node \(v_i\), we extract an induced subgraph \(G_i = (\mathbf{P}_i, \mathbf{X}_i, \mathbf{A}_i)\). The node features \(\mathbf{X}_i\) and adjacency \(\mathbf{A}_i\) are obtained either from the \(k\)-hop neighborhood centered at \(v_i\) or from subgraphs sampled via a random-walk procedure, while \(\mathbf{P}_i\) denotes positional encodings. This subgraph is then processed by the graph encoder \(g_{\text{victim}}\), such as GraphGPS~\cite{rampavsek2022recipe}, to capture its structural and semantic information:
\begin{equation}
    g_{\text{victim}}(v_i) = P(g_{\text{victim}}(G_i)). \label{eq:graph-encoding}
\end{equation}
A readout operation (e.g., mean pooling) $P$ is applied over node representations to obtain a fixed-size embedding \(g_{\text{victim}}(v_i) \in \mathbb{R}^{1 \times d}\), serving as the representation of node \(v_i\) in the latent space.

\noindent \textbf{Text Encoder.}
During training, GraphCLIP leverages graph-text pairs, where each node (or subgraph) is associated with a textual summary that describes its semantics, structure, or context.
These graph summaries are collected from diverse graph datasets across multiple domains (e.g., citation networks, e-commerce, and social graphs) and serve as natural-language descriptions of graph contents.
Each textual summary \(T_i\) is encoded into a dense embedding using a pretrained text encoder \(h_\phi\) (e.g., MiniLM \cite{wang2020minilm}):
\begin{equation}
    Z_i = h_\phi(T_i), \label{eq:text-encoding}
\end{equation}
where \(Z_i \in \mathbb{R}^{1 \times d}\) represents the text embedding corresponding to the $i$-th graph or node instance.


\noindent \textbf{Contrastive Alignment.}
Given the paired graph embeddings \(\{g_{\text{victim}}(v_i)\}\) and text embeddings \(\{Z_i\}\), GraphCLIP is trained using a contrastive objective \cite{oord2018representation} to align the two modalities in a shared latent space.
Specifically, the model encourages matched graph-text pairs to have high cosine similarity while pushing apart unmatched pairs:
\begin{equation}
    \mathcal{L}_{\text{con}} = - \frac{1}{N} \sum_{i=1}^{N}
    \log \frac{\exp(\text{sim}(g_{\text{victim}}(v_i), Z_i)/\tau)}
    {\sum_{j=1}^{N} \exp(\text{sim}(g_{\text{victim}}(v_i), Z_j)/\tau)}, \label{eq:contrastive-loss}
\end{equation}
where \(\text{sim}(\cdot,\cdot)\) denotes cosine similarity and \(\tau\) is a temperature parameter. Through large-scale pretraining over diverse graph-text pairs, this objective establishes cross-modal correspondence between graph structures and their textual summaries.

\subsection{Zero-Shot Inference with GFM}
After training, given any node from an unseen graph, the GFM performs zero-shot classification by computing similarity between the node embedding produced by the graph encoder and the semantically meaningful label embeddings generated by the text encoder:
\begin{equation}
\hat{y}_i \;=\;
\underset{k\in\{1,\dots,K\}}{\operatorname*{arg\,max}}
\operatorname{sim}\!\bigl(g_{\text{victim}}(v_i), Z_k\bigr),
\end{equation}
where \(\operatorname{sim}(g_{\text{victim}}(v_i), Z_k)\) denotes the similarity between the node embedding \(g_{\text{victim}}(v_i)\) and the text embedding \(Z_k\) corresponding to the \(k\)-th label. Here, a \emph{label sentence} is a short natural-language phrase describing a class (e.g., ``a paper about deep learning’’), and the sentence with the highest similarity determines the predicted label of node \(v_i\).

\begin{figure*}[!t]
   \begin{center}
   \includegraphics[width=.95\linewidth]{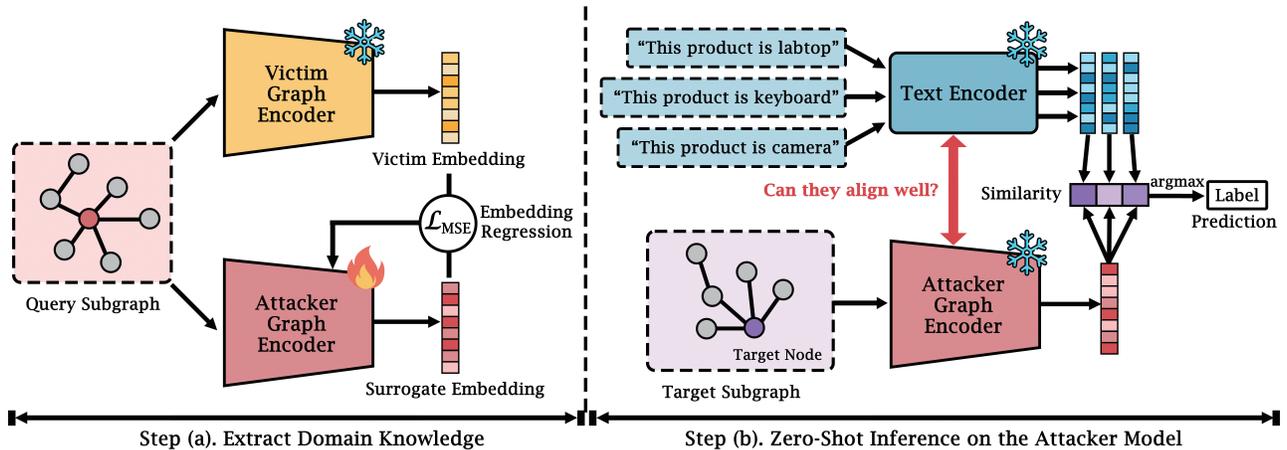} %
    \end{center}
 \caption{MEA against a GFM. (a) Extracting domain knowledge: the attacker queries the victim graph encoder with training graphs and uses the returned embeddings as supervision to train an attacker graph encoder.
(b) Zero-shot inference using the attacker model: after training, the attacker combines its encoder with the frozen text encoder and performs zero-shot inference on unseen graphs, showing that the extracted model inherits the victim’s semantic alignment and inference capabilities.}
\label{fig:MEA_GFM}
\end{figure*}
\section{Threat Model and Attack Taxonomy}
\label{sec:threat}

\subsection{Attacker Setting}
\noindent\textbf{Graph-summary/caption pretraining data.}
Unlike the image domain, where large-scale image-caption datasets are readily obtainable online, high-quality graph-summary or caption data are typically scarce and are often generated by the model owner. As a result, they are private and costly to create.
In this work, we assume that the attacker has no access to the graph-summary or caption data used for pretraining the victim model. 

\noindent\textbf{Victim model.}
Following standard MEA assumptions, the attacker does not have access to the victim graph encoder's parameters. 
However, the attacker can query the victim model and obtain the output embeddings \( g_\text{victim}(v_i) \in \mathbb{R}^d \) produced by its graph encoder. 
The attacker may or may not know the architectural family of the victim encoder (e.g., Transformer-style vs.\ GNN-style). 
Since the text encoder in GraphCLIP remains frozen during pretraining, it typically employs an open-source pretrained language model; thus, in this work, the attacker uses the same text encoder as the victim model.

\noindent\textbf{Attributed graph datasets.}
Because numerous open-source attributed graph datasets are publicly available, we assume that the attacker can access such datasets for querying. 
However, these datasets do not include the proprietary graph-summary or caption data used by the GFM owner.

\noindent\textbf{Query budget.}
We consider both unlimited and limited query budgets \(B\). 
Queries can be drawn from one or multiple \emph{source} TAG datasets or constructed in a \emph{target-only} manner—i.e., sampled exclusively from the target graph.

\subsection{Attacker Goal}
The objective of a MEA is to recover the functional behavior of a target model by training a surrogate that produces similar outputs and downstream performance. 
Existing MEA work on graph models has focused primarily on small GNNs and therefore targets only \emph{graph-specific} extraction: the surrogate is optimized to mimic the victim on a \emph{single} graph and does not generalize across graphs or domains.
GFMs significantly broaden the attack surface. Because they are pretrained on diverse datasets and pair a graph encoder with a frozen text encoder, they support zero-shot inference on many unseen graphs. Extracting such a model therefore allows an attacker to recover \emph{domain-level} knowledge—capturing patterns that apply across multiple graphs within a domain, not just one.
As GFMs grow in scope and capability, extracting their functional behavior becomes more consequential: a successful surrogate can inherit domain-wide zero-shot inference ability, raising stronger security and IP concerns.  
We study two attacker objectives in this work:
\begin{itemize}[leftmargin=*]
  \item \textbf{Full-model / domain-level extraction (G1).} Learn a surrogate encoder that captures the victim’s domain-level knowledge, generalizes across many graphs, and supports zero-shot inference on new, unseen graphs.
  \item \textbf{Graph-specific extraction (G2).} Learn a surrogate encoder that matches the victim’s behavior on one target graph, without aiming for broader cross-graph generalization.
\end{itemize}

\subsection{Attack Taxonomy}
\label{subsec:taxonomy}
In all cases the attacker has no access to the victim graph encoder's parameters nor to the proprietary graph-summary pretraining data. Attack scenarios vary along four axes: (i) extraction goal (cross-domain vs.\ single-graph), (ii) surrogate capacity / architectural knowledge, (iii) query budget, and (iv) data availability (complete vs.\ partial or mismatched). We consider six practical attack scenarios; due to space limitations, a visual illustration is provided in Appendix~\ref{appen:taxonomy}.
\begin{itemize}[leftmargin=*]
\item \textbf{Attack-1 (Full-Model Extraction).}   
The attacker aims to extract the victim GFM’s full functionality across all domains encoded in the model. The attacker may or may not know the victim encoder’s architectural family; when known they can design a matched-capacity surrogate, and when unknown they train a mismatched or lower-capacity surrogate. The attacker has access to pretraining attributed-graph datasets covering the model’s domains and an unrestricted query budget. 

\item \textbf{Attack-2 (Domain-Specific
Extraction).}  
The attacker targets a single domain (e.g., academic, e-commerce, social). 
With unrestricted query access to public pretraining datasets from the domain, the attacker gathers victim-model embeddings to train a surrogate encoder that preserves domain-level knowledge and enables zero-shot inference on unseen graphs within that domain.

  \item \textbf{Attack-3 (Budget-Constrained Extraction).} 
  The attacker has the same domain-level objective but is limited to a finite query budget B. They must strategically select informative subgraphs to query the victim model in order to maximize coverage and sample efficiency.

  \item \textbf{Attack-4 (Graph-Specific Extraction).} 
  The goal is to reproduce the victim’s behavior on a single target graph, without requiring cross-graph generalization. The attacker has no access to external pretraining datasets and therefore constructs all queries solely from that graph.

    \item \textbf{Attack-5 (Extraction with Synthetic Graphs).}
For all training attributed-graph datasets, the attacker knows only the features of a small subset of nodes and their local neighborhoods.  
To query the victim model, the attacker must synthesize missing node attributes to assemble a queryable subgraph.  
Evaluation measures how much full-model knowledge can be extracted under this more realistic partial-access constraint.

    \item \textbf{Attack-6 (Training-Data-Free Extraction).}
The attacker does not know which attributed graph datasets the victim GFM used for training. To perform full model extraction, the attacker therefore relies on other available public graph datasets to query the victim model.

\end{itemize}

\section{MEA-GFM Attack}
\label{sec:method}

\subsection{MEA on the Graph Encoder}
As discussed above, in many multimodal systems the text encoder is an open-source, pretrained language model while the modality-specific encoders (e.g., vision, graph) are trained by the model owner to capture modality-specific semantics and to align with the text encoder. In this work, we therefore focus on extracting the knowledge embedded in the victim’s multimodal \emph{graph encoder} and training a surrogate that replicates its function. The general attack process is shown in Fig. \ref{fig:MEA_GFM}.

Concretely, the attacker first chooses a surrogate graph encoder architecture $g_{\text{attack}}$ (e.g., GCN, GAT, or a lightweight graph transformer) and then minimizes an embedding regression objective against the victim graph encoder $g_{\text{victim}}$:
\begin{equation}
\mathcal{L}_{\mathrm{MSE}}
\;=\;
\sum_{i=1}^{N} \big\| g_{\text{attack}}(v_i) - g_{\text{victim}}(v_i) \big\|_2^2,
\label{eq:mse}
\end{equation}
where $N$ is the number of queried training subgraphs. In practice, embeddings from both encoders are $\ell_2$-normalized before computing the loss or cosine similarity, and gradients are not propagated through $g_{\text{victim}}$.

\noindent \textbf{Alignment challenge.}
Minimizing \eqref{eq:mse} aligns the surrogate to the victim’s graph embedding space, but does not by itself guarantee that the surrogate will be \emph{aligned to the text encoder} used at inference. This alignment is essential: only when the surrogate embeddings are compatible with the text encoder’s representation can the attacker perform reliable zero-shot retrieval or classification.

\noindent \textbf{Solutions to ensure text-graph alignment.}
In the vision domain, many works combine embedding regression with a contrastive loss to produce a surrogate that is both a faithful replica of the victim encoder and well aligned with the text encoder. When paired graph-text examples $(v_i, T_i)$ and a compatible public text encoder $h_\phi$ are available, one can add a CLIP-style contrastive term \cite{radford2021learning}:
\begin{equation}
    \mathcal{L}_{\mathrm{contrast}}
    = -\frac{1}{|\mathcal{B}|}\sum_{i\in\mathcal{B}}
    \log \frac{\exp\big(\langle g_{\text{attack}}(v_i), Z_i\rangle/\tau\big)}
    {\sum_{j\in\mathcal{B}}\exp\big(\langle g_{\text{attack}}(v_i), Z_j\rangle/\tau\big)},
    \label{eq:contrastive}
\end{equation}
where $Z_i = h_\phi(T_i)$ is the text embedding of the summary data associated with $v_i$, $\tau$ is a temperature, and $\mathcal{B}$ is the minibatch. A combined objective
\begin{equation}
    \mathcal{L} = \lambda_{\mathrm{MSE}}\mathcal{L}_{\mathrm{MSE}}
         + \lambda_{\mathrm{contrast}}\mathcal{L}_{\mathrm{contrast}}
\end{equation}
where $\mathcal{L}_{\mathrm{MSE}}$ matches embeddings and $\mathcal{L}_{\mathrm{contrast}}$ preserves graph-text alignment, allowing the surrogate to recover both geometric and multimodal behavior of the victim model.

\noindent \textbf{Challenges of contrastive learning for MEA.}
However, in many real-world MEA scenarios, obtaining high-quality paired graph-text examples is difficult or impossible. 
Moreover, contrastive pretraining is computationally expensive: it requires large batches and more training steps, resulting in significantly higher cost than supervised learning. 

\noindent \textbf{MEA without access to graph-summary data.}  
In this work, we assume the attacker has no access to private graph-summary (or caption) data and therefore adopts a single supervised learning loss as the \emph{only} training objective. 
Concretely, the attacker selects $g_{\text{attack}}$ and minimizes the direct regression loss against $g_{\text{victim}}$ in \eqref{eq:mse}. 
We next demonstrate that this alone allows the attacker’s graph encoder to remain aligned with the frozen text encoder and thereby inherit the victim model’s zero-shot inference capability.

\subsection{Attacker Graph Encoder-Text Encoder Alignment}
We now show that, under mild assumptions, minimizing $\mathcal{L}_{\mathrm{MSE}}$ alone is sufficient for the surrogate to preserve the victim's zero-shot node classification ability. In practice, the attacker queries $g_{\text{victim}}$ on a collection of training graphs, learns $g_{\text{attack}}$ to approximate those responses, and then performs zero-shot inference on entirely \emph{unseen} graphs from the same domain. Intuitively, if $g_{\text{attack}}$ closely matches $g_{\text{victim}}$ on the queried graphs, and if queried nodes are representative enough that every unseen test node is close to at least one of them in embedding space, then the surrogate encoder will behave similarly to the victim on new graphs when paired with the same text encoder.

\vspace{0.6em}
\noindent\textbf{Lemma 1.}
If the surrogate encoder produces node embeddings that are close to the victim’s embeddings on the queried nodes, and if every unseen test node is close (in embedding space) to at least one queried node, then the surrogate will make nearly identical zero-shot decisions as the victim. Formally, Lemma~1 relies on the following assumptions.

\begin{enumerate}[leftmargin=*, itemindent=0pt]
    \item \textbf{Training coverage.}  
    For every unseen test node $v \in V_{\text{test}}$, there exists a queried node $\tilde{v} \in V_{\text{train}}$ such that
    \begin{equation}
        \| g_{\text{victim}}(v) - g_{\text{victim}}(\tilde{v}) \|_2 < \varepsilon,
        \\
        \| g_{\text{attack}}(v) - g_{\text{attack}}(\tilde{v}) \|_2 < \varepsilon.
        \label{eq:assump1}
    \end{equation}
    This means that for every unseen node, there exists a queried node that lies close to it under both encoders.

    \item \textbf{Embedding alignment on queried nodes.}  
    For every queried node $\tilde{v} \in V_{\text{train}}$,
    \begin{equation}
        \| g_{\text{victim}}(\tilde{v}) - g_{\text{attack}}(\tilde{v}) \|_2 < \varepsilon.
        \label{eq:assump2}
    \end{equation}
This states that the surrogate matches the victim’s embeddings on every queried node.
\end{enumerate}

Let
\begin{equation}
\begin{aligned}
c_{\text{victim}}(v) &= \arg\max_{k} \langle g_{\text{victim}}(v), Z_k \rangle, \\
c_{\text{attack}}(v) &= \arg\max_{k} \langle g_{\text{attack}}(v), Z_k \rangle,
\end{aligned}
\end{equation}
denote the zero-shot predictions made by the victim and surrogate. Define 
\begin{equation}
    \kappa := \| Z_{c_{\text{attack}}(v)} - Z_{c_{\text{victim}}(v)} \|_2 .
\end{equation}
Here, $Z_{c_{\text{attack}}(v)}$ and $Z_{c_{\text{victim}}(v)}$ denote the text embeddings of the labels predicted by the attacker and the victim, respectively. 
With unit-normalized text embeddings, we have $\kappa \le 2$.  
Then for every unseen test node $v$,
\begin{equation}
    \langle g_{\text{attack}}(v), Z_{c_{\text{attack}}(v)} \rangle
    -
    \langle g_{\text{attack}}(v), Z_{c_{\text{victim}}(v)} \rangle
    < 3\,\kappa\,\varepsilon \;\le\; 6\varepsilon .
    \label{eq:lemma_bound_kappa}
\end{equation}

The proof of Lemma 1 is in Appendix \ref{appen:proof}. The inequality in \eqref{eq:lemma_bound_kappa} bounds how much the surrogate’s similarity score for its chosen label can exceed the similarity score assigned to the victim’s chosen label. Since zero-shot predictions are made by selecting the label
$\arg\max_{k} \langle g(v), Z_k \rangle$, a small gap in~\eqref{eq:lemma_bound_kappa} implies that
\begin{equation}
    c_{\text{attack}}(v) = c_{\text{victim}}(v).
\end{equation}
whenever $3\,\kappa\,\varepsilon$ is sufficiently small. Empirically, after optimizing $\mathcal{L}_{\mathrm{MSE}}$, the observed embedding discrepancy satisfies
\begin{equation}
    \| g_{\text{victim}}(v) - g_{\text{attack}}(v) \|_2 \approx 0,
\end{equation}
leading to near-perfect agreement on entirely unseen graphs, without requiring any additional labels or graph-text pairs on those unseen graphs.

\vspace{0.4em}
\noindent\textbf{Interpretation and Takeaway.}
Lemma~1 provides a sufficient condition for zero-shot inference consistency between the attacker model and the victim model: if the surrogate matches the victim on the queried nodes, and if those queried nodes sufficiently cover the embedding space, then the surrogate will inherit the victim’s zero-shot decision behavior when paired with the same text encoder. Consequently, a simple $\ell_2$ regression loss on node embeddings is enough to replicate the victim’s zero-shot inference capability. Once $g_{\text{attack}}$ approximates $g_{\text{victim}}$ on the training quries, the surrogate remains aligned with the text encoder and preserves the victim’s generalization to entirely unseen nodes or subgraphs—without requiring graph-text pairs or contrastive learning.

\vspace{0.4em}
\noindent\textbf{Applicability to other multimodal foundation models.}
Although we focus on GraphCLIP, our method is not specific to the graph domain. Any multimodal foundation model that follows a similar architecture, namely a modality-specific encoder aligned with a public text encoder, can be attacked using the same embedding regression strategy. This includes vision-language, audio-language, and point cloud-language models built on CLIP-style alignment. Therefore, the proposed approach provides a generic pathway for extracting zero-shot capability from a broad class of multimodal foundation models.

\subsection{Attack Methods}
\label{subsec:attack methods}

Across all attacks, the attacker submits a collection of graphs or induced subgraphs \(\{G_i\}\) to the victim API and receives the corresponding graph embeddings $\{g_{\text{victim}}(G_i) \in \mathbb{R}^d\}$.
The attacker then trains a surrogate encoder \(g_{\text{attack}}(\cdot)\) to regress onto these returned embeddings.

\subsubsection{Full-Model Extraction} 
The attacker aims to extract the victim GFM’s full functionality across all domains encoded in the model by training a surrogate graph encoder. The attacker may or may not know the victim encoder’s architectural family; when architecture is known they can design a matched-capacity surrogate, and when unknown they train a mismatched or lower-capacity surrogate. We assume an unrestricted query budget: the attacker can construct arbitrarily many query subgraphs sampled from multiple public TAG  pretraining datasets and submit them to the victim graph encoder to collect embeddings. These returned embeddings provide supervision for training the surrogate via supervised regression (Eq.~\eqref{eq:mse}). Because the victim’s text encoder is typically public and frozen, Lemma~1 implies that a surrogate which closely matches the victim embeddings on representative training data will be aligned with the text encoder and thus support zero-shot inference on previously unseen graphs. With abundant queries and broad coverage, this attack is expected to produce a high-fidelity surrogate that replicates the victim’s zero-shot decision behavior on any downstream graph the victim can handle.

\subsubsection{Domain-Specific Extraction}
The attacker focuses on extracting knowledge from a \emph{single} domain (e.g., academic, e-commerce, social), aiming for a surrogate that generalizes across graphs in that domain and supports domain-specific zero-shot inference. As in Attack-1, the attacker may or may not know the victim’s architecture; when unknown, they train a mismatched or lower-capacity encoder. The attacker queries all pretraining datasets from the target domain, obtains graph embeddings from the victim, and trains the surrogate using supervised regression.  
This setting evaluates how well domain-level knowledge can be transferred when the surrogate architecture is simpler or mismatched. We measure embedding fidelity and zero-shot accuracy on unseen graphs within the same domain, testing whether lightweight GNNs (e.g., GCN, GAT) can emulate a large graph transformer pretrained with graph-text contrastive learning.

\subsubsection{Budget-Constrained Extraction} 
This attack follows the goal of domain-level extraction, but the attacker is limited to a finite query budget \(B\). 
Consequently, the attacker must strategically select a subset of nodes or induced subgraphs to query the victim model and use the returned embeddings to train the surrogate encoder. 
We evaluate how different budget sizes and query selection policies affect surrogate fidelity and zero-shot performance on unseen graphs.

\subsubsection{Graph-Specific Extraction} 
This attack models the classical graph-specific extraction scenario: the attacker has no access to external (pretraining) datasets and must construct all queries from the target graph itself. 
The attacker construct queries sampled from the target graph, obtains the victim's output embeddings for those queries, and trains a surrogate encoder to reproduce the victim's behavior on that particular graph (graph-specific extraction, G2). 
Because the objective is confined to a single graph, cross-graph generalization is not required. 
This setting is realistic for adversaries whose goal is to replicate a provider's service on one target graph; evaluation therefore focuses on inference fidelity and downstream performance restricted to that graph.

\subsubsection{Extraction with Synthetic Graphs}
In this setting the attacker aims for full-model extraction but only knows a small portion of each training graph. 
Specifically, the overall procedure is:

\begin{enumerate}[leftmargin=*]
  \item \textbf{Visibility constraint:} The attacker only has knowledge of a small subset of nodes and the local structure around them: specifically, the attacker knows the features of the nodes they control and the edges connecting those nodes to their 1-hop (and possibly 2-hop) neighbors, but does not have access to the features of the neighboring nodes or any other nodes in the graph.
  For example, an adversary who gains access to a few bank or social-network accounts can obtain those users’ profile information and see their transaction or friend connections, yet the attributes of the connected accounts remain unknown.
\item \textbf{Attribute synthesis:} To obtain a valid query graph, the attacker synthesizes missing features for a target node \(v\) by averaging the known features of its 1-hop and 2-hop neighbors, assuming nearby nodes tend to have similar attributes.
For an unknown node \(v\), its synthesized feature is 
\begin{equation}
    x'_{v}=
    \alpha \cdot \frac{1}{|N(v)|}\sum_{u\in N(v)} x_u
    + (1-\alpha) \cdot \frac{1}{|N^2(v)|}\sum_{u\in N^2(v)} x_u,
\end{equation}
where \(N(v)\) and \(N^2(v)\) denote the 1-hop and 2-hop neighbors of \(v\), respectively, and \(\alpha\in[0,1]\) balances their contributions.

  \item \textbf{Subgraph construction and querying:} After imputing missing attributes for target nodes, the attacker assembles complete synthetic subgraphs \(G_i\) and queries the victim GFM to obtain graph embeddings \(g_{\mathrm{victim}}(G_i)\in\mathbb{R}^d\).
  \item \textbf{Attacker model training:} Using a collection of synthetic subgraphs and their corresponding victim embeddings, the attacker trains a surrogate graph encoder with the supervised loss \(\mathcal{L}_{\mathrm{MSE}}\).
  \item \textbf{Goal:} Assess whether a surrogate can still extract the full functionality of the victim model, even when only partial node features and local connectivity are available.
\end{enumerate}

This procedure models realistic adversaries who control or inject a small number of accounts (e.g., new users) and lack global visibility. By synthesizing plausible neighbor features, the attacker attempts to reconstruct a high-fidelity, cross-domain surrogate despite limited per-query information.

\subsubsection{Training-Data-Free Extraction}
In this attack the attacker attempts full-model extraction but does not know which attributed-graph datasets the victim GFM used for pretraining. To approximate the victim’s full functionality, the attacker therefore collects alternative public attributed graph datasets and uses them to construct query subgraphs. These substitute datasets may differ in distribution from the victim’s original pretraining corpus, resulting in training data discrepancies. Evaluation measures how this data mismatch affects the fidelity of the extracted surrogate and its zero-shot performance on entirely unseen graphs. 

\section{Evaluation}
\subsection{Experimental Setup}
\label{subsec:exp_setup}
\noindent \textbf{Datasets.} The datasets used in our experiments are listed in Table~\ref{tab:datasets}. 
Both the pretraining datasets and the evaluation datasets follow the same setup as GraphCLIP~\cite{zhu2025graphclip}. 
In addition, for the Attack-6 setting we include three extra datasets (\emph{Books-Children}, \emph{Sports-Fitness}, and \emph{DBLP}) to construct query subgraphs and train the attacker models.
For each graph in the evaluation datasets listed in Table~\ref{tab:datasets}, we split the nodes into a 60\% training set, 10\% validation set, and 30\% test set.

\noindent \textbf{Models.} In our experiments the target model is the released GraphCLIP checkpoint \cite{zhu2025graphclip}, which comprises 12 GPS Graph Transformer layers \cite{rampavsek2022recipe} and 128,007,618 trainable parameters. 
For attacker models we use three configurations. 
The first is a GraphGPS~\cite{rampavsek2022recipe} with reduced depth and hidden dimension; relative to the GPS encoder in GraphCLIP this attacker has 11,102,400 trainable parameters. 
The second is a three-layer GAT \cite{velivckovic2017graph} with 2,931,456 trainable parameters. 
The third is a three-layer GCN \cite{kipf2016semi} with 952,064 trainable parameters.
All attackers are trained with the AdamW optimizer \cite{loshchilov2019decoupledweightdecayregularization}, using a learning rate of 0.0001 and a weight decay of 0.00001. A batch size of 32 is used. All experiments are conducted using only one NVIDIA RTX 6000 Ada GPU.

\begin{table}[t]
\centering
\caption{Statistics of the pretraining and evaluation TAG datasets. \#C denotes the number of classes.}
\resizebox{\columnwidth}{!}{
\begin{tabular}{lcccr}
\toprule
\textbf{Dataset} & \textbf{\#Nodes} & \textbf{\#Edges} & \textbf{\#C} & \textbf{Domain} \\
\midrule
\multicolumn{5}{c}{\textit{Pretraining Datasets}} \\
\midrule
OGBN-ArXiv \cite{wang2020microsoft}       & 169,343   & 1,166,243   & 40 & Academic \\
ArXiv-2023 \cite{he2023harnessing}      & 46,198    & 78,543      & 40 & Academic \\
PubMed  \cite{sen2008collective}         & 19,717    & 44,338      & 3  & Academic \\
OGBN-Products \cite{hu2020open}    & 2,449,029 & 61,859,140  & 47 & E-commerce \\
Reddit \cite{huang2024can}          & 33,434    & 198,448     & 2  & Social \\
\midrule
\multicolumn{5}{c}{\textit{Evaluation Datasets}} \\
\midrule
Cora \cite{mccallum2000automating}            & 2,708   & 5,429    & 7  & Academic \\
CiteSeer \cite{giles1998citeseer}        & 3,186   & 4,277    & 6  & Academic \\
Ele-Photo \cite{yan2023comprehensive}       & 8,432   & 500,928  & 10 & E-commerce \\
Ele-Computers \cite{yan2023comprehensive}    & 87,229  & 721,081  & 10 & E-commerce \\
Books-History \cite{yan2023comprehensive}   & 51,551  & 358,574  & 12 & E-commerce \\
WikiCS \cite{mernyei2020wiki}          & 11,701  & 215,863  & 10 & Wikipedia \\
Instagram \cite{huang2024can}      & 11,339  & 144,010  & 9  & Social \\
\midrule
\multicolumn{5}{c}{\textit{Extra Query Datasets}} \\
\midrule
Books-Children \cite{yan2023comprehensive}   & 76,875  & 1,554,578  & 24 & E-commerce \\
Sports-Fitness \cite{yan2023comprehensive}   & 173,055  & 1,773,500  & 13 & E-commerce \\
DBLP \cite{chen2024text}  & 14,376  & 431,326  & 4 & Academic \\

\bottomrule
\end{tabular}}
\vspace{-0.1in}
\label{tab:datasets}
\end{table}

\noindent\textbf{Evaluation Metrics.} Let $\mathcal{D}=\{(x_i,y_i)\}_{i=1}^{N}$ be the evaluation set with ground-truth labels $y_i$, let $A(x)$ denote the attacker’s prediction, and let $V(x)$ denote the victim model’s prediction. Using the indicator $\mathbf{1}\{\cdot\}$, the metrics are defined independently as
\begin{equation}
\mathrm{Accuracy}(A;\mathcal{D}) \;=\; \frac{1}{N}\sum_{i=1}^{N} \mathbf{1}\!\big\{A(x_i)=y_i\big\},
\end{equation}
\begin{equation}
\mathrm{Fidelity}(A,V;\mathcal{D}) \;=\; \frac{1}{N}\sum_{i=1}^{N} \mathbf{1}\!\big\{A(x_i)=V(x_i)\big\}.
\end{equation}

\begin{table*}[t]
  \caption{Use all pretraining datasets to perform MEA and extract the victim GFM's \emph{full} knowledge and \emph{full} functionality.
  }
  \label{tab:full_domain}
  \centering
  \renewcommand{\arraystretch}{1.1}
  \setlength{\tabcolsep}{4pt}
  \resizebox{\textwidth}{!}{
  \begin{tabular}{c|*{7}{cc}}
    \hline
    \multirow{2}{*}{\textbf{Model}} &
    \multicolumn{2}{c}{\textbf{Cora}} &
    \multicolumn{2}{c}{\textbf{CiteSeer}} &
    \multicolumn{2}{c}{\textbf{WikiCS}} &
    \multicolumn{2}{c}{\textbf{Instagram}} &
    \multicolumn{2}{c}{\textbf{Ele-Photo}} &
    \multicolumn{2}{c}{\textbf{Ele-Computers}} &
    \multicolumn{2}{c}{\textbf{Books-History}} \\
    \cline{2-15}
    & \textbf{Acc.} & \textbf{Fid.}
    & \textbf{Acc.} & \textbf{Fid.}
    & \textbf{Acc.} & \textbf{Fid.}
    & \textbf{Acc.} & \textbf{Fid.}
    & \textbf{Acc.} & \textbf{Fid.}
    & \textbf{Acc.} & \textbf{Fid.}
    & \textbf{Acc.} & \textbf{Fid.} \\
    \hline
    GPS
    & 57.69 & 82.53
    & 66.53 & \textbf{94.25}
    & 68.50 & \textbf{89.92}
    & \textbf{59.70} & \textbf{93.77}
    & 45.37 & \textbf{74.84}
    & 56.12 & 75.59
    & 52.82 & \textbf{90.00} \\
    GAT
    & \textbf{68.51} & \textbf{83.15}
    & \textbf{68.41} & 90.79
    & \textbf{70.38} & 85.82
    & 58.44 & 83.04
    & \textbf{55.01} & 74.38
    & \textbf{65.25} & \textbf{79.73}
    & 47.99 & 83.80 \\
    GCN
    & 68.14 & 81.55
    & 67.68 & 89.96
    & 68.84 & 80.43
    & 57.61 & 81.28
    & 52.33 & 68.07
    & 64.33 & 76.92
    & \textbf{53.51} & 82.91 \\
    \hline
    Victim
    & 64.94 & NA
    & 68.83 & NA
    & 70.09 & NA
    & 61.29 & NA
    & 54.13 & NA
    & 64.46 & NA
    & 53.32 & NA \\
    \hline
  \end{tabular}
  }
\end{table*}

\begin{table*}[t]
  \caption{For each domain, we perform MEA using only that domain’s training data to extract domain-specific knowledge from the victim GFM and evaluate the resulting surrogate on all evaluation graphs.
  All values are percentages (\%).
  }
  \label{tab:domain_specific}
  \centering
  \renewcommand{\arraystretch}{1.1}
  \setlength{\tabcolsep}{4pt}
  \resizebox{\textwidth}{!}{
  \begin{tabular}{c|c|*{7}{cc}}
    \hline
    \multirow{2}{*}{\textbf{Pretrain Domain}} &
    \multirow{2}{*}{\textbf{Model}} &
    \multicolumn{2}{c}{\textbf{Cora}} &
    \multicolumn{2}{c}{\textbf{CiteSeer}} &
    \multicolumn{2}{c}{\textbf{WikiCS}} &
    \multicolumn{2}{c}{\textbf{Instagram}} &
    \multicolumn{2}{c}{\textbf{Ele-Photo}} &
    \multicolumn{2}{c}{\textbf{Ele-Computers}} &
    \multicolumn{2}{c}{\textbf{Books-History}} \\
    \cline{3-16}
    & &
    \textbf{Acc.} & \textbf{Fid.} &
    \textbf{Acc.} & \textbf{Fid.} &
    \textbf{Acc.} & \textbf{Fid.} &
    \textbf{Acc.} & \textbf{Fid.} &
    \textbf{Acc.} & \textbf{Fid.} &
    \textbf{Acc.} & \textbf{Fid.} &
    \textbf{Acc.} & \textbf{Fid.} \\
    \hline
    \multirow{3}{*}{\centering \textbf{Academic}} & GPS
    & 58.43 & 81.67
    & 66.21 & 89.54
    & 67.96 & \textbf{89.03}
    & 60.46 & 91.95
    & 44.66 & 65.68
    & 52.95 & 67.30
    & 47.16 & 77.33 \\
    & GAT
    & \textbf{67.77} & 83.27
    & \textbf{68.41} & 90.27
    & \textbf{70.83} & 83.94
    & 59.47 & 80.07
    & 47.92 & 56.82
    & 58.53 & 66.66
    & 39.76 & 71.01 \\
    & GCN
    & 67.65 & 82.66
    & 67.78 & \textbf{90.59}
    & 69.87 & 79.27
    & 56.00 & 70.37
    & 45.33 & 54.52
    & 53.94 & 60.06
    & 36.37 & 65.39 \\
    \hdashline
    \multirow{3}{*}{\centering \textbf{Social}} & GPS
    & 54.00 & 60.52
    & 59.83 & 77.51
    & 61.58 & 76.64
    & 63.23 & \textbf{95.77}
    & 48.28 & 73.22
    & 58.45 & 73.43
    & 54.17 & 80.12 \\
    & GAT
    & 57.32 & 65.68
    & 59.00 & 70.50
    & 63.60 & 75.25
    & \textbf{63.52} & 93.42
    & 48.67 & 73.07
    & 40.26 & 50.19
    & \textbf{58.95} & 73.35 \\
    & GCN
    & 33.95 & 36.53
    & 36.19 & 42.47
    & 52.35 & 58.39
    & 63.29 & 93.30
    & 38.95 & 56.94
    & 31.51 & 39.29
    & 57.28 & 67.34 \\
    \hdashline
    \multirow{3}{*}{\centering \textbf{E-commerce}} & GPS
    & 63.10 & \textbf{85.24}
    & 65.17 & 89.64
    & 68.70 & 88.44
    & 55.47 & 69.84
    & 44.29 & 71.08
    & 54.14 & 70.57
    & 51.92 & \textbf{90.24} \\
    & GAT
    & 66.67 & 81.30
    & 65.79 & 87.03
    & 65.91 & 80.18
    & 57.82 & 80.54
    & \textbf{54.25} & \textbf{78.30}
    & \textbf{63.88} & \textbf{79.60}
    & 50.14 & 85.67 \\
    & GCN
    & 65.19 & 77.98
    & 64.44 & 85.67
    & 65.05 & 76.76
    & 58.38 & 81.45
    & 54.24 & 74.31
    & 62.28 & 74.16
    & 52.89 & 84.18 \\
    \hline
    \textbf{All}
    & Victim
    & 64.94 & NA
    & 68.83 & NA
    & 70.09 & NA
    & 61.29 & NA
    & 54.13 & NA
    & 64.46 & NA
    & 53.32 & NA \\
    \hline
  \end{tabular}
  }
\end{table*}

\noindent \textbf{Detailed configurations of the six attacks.}

\noindent \textbf{Attack-1.} We conduct a full-corpus extraction attack using all pretraining datasets. Three attacker architectures—GPS, GAT, and GCN—are trained for 2 epochs on subgraph-embedding pairs obtained by querying the victim model across all pretraining datasets. Each trained attacker is then evaluated on the test split of every evaluation dataset.

\noindent \textbf{Attack-2.}  We again employ all pretraining datasets to perform the extraction attack. For each domain—Academic, Social, and E-commerce—we train domain-specific attackers on the extracted subgraph-embedding pairs for 8 epochs, using the three attacker architectures (GPS, GAT, GCN). This yields nine attacker models in total (3 domains $\times$ 3 architectures). Each trained model is evaluated on the test split of every evaluation dataset.

\noindent \textbf{Attack-3.} We investigate MEA under a query budget within the academic domain, using budgets \{1{,}000, 500, 250, 100\}. For each budget, we consider two sampling schemes: (i) \emph{single-dataset sampling}—randomly selecting budget-sized subgraphs from one of OGBN-ArXiv, ArXiv-2023, or PubMed; and (ii) \emph{mixed-dataset sampling}—drawing 30\%, 30\%, and 40\% budget-sized queries from  OGBN-ArXiv, ArXiv-2023, and PubMed, respectively. Attackers are trained for \{30, 60, 120, 200\} epochs on the embeddings returned by the victim model under the specified budget and evaluated on the academic-domain evaluation datasets.

\noindent \textbf{Attack-4.} We examine a same-graph generalization setting. For each evaluation dataset, we first extract 100 subgraph-embedding pairs from its training split and train the attacker models for 60 epochs on these pairs. Evaluation is then conducted on the testing split of the \emph{same} graph.

\noindent \textbf{Attack-5.} To impose the visibility constraint, we randomly sample 10\% of nodes from OGBN-ArXiv, ArXiv-2023, PubMed, and Reddit, and 1\% from OGBN-Products as the visible subset (i.e., nodes whose attributes are known to the attacker). We then construct the synthetic graphs following the procedure described in Section~\ref{subsec:attack methods}. All three attacker models are trained for 2 epochs on embeddings extracted from all synthetic graphs and are subsequently evaluated on the test split of the evaluation datasets.

\noindent \textbf{Attack-6.}  We train the attacker models on embeddings extracted from the extra query datasets listed in Table~\ref{tab:datasets} for 4 epochs. The resulting models are evaluated on the test split of all evaluation datasets.

\subsection{Evaluation Results}
\subsubsection{Replicating the Whole GFM with a Single Loss}
We begin by evaluating whether a single supervised MSE loss (Eqn.~\ref{eq:mse}) is sufficient to replicate a large GFM. The attacker queries the victim’s graph encoder on all pretraining datasets listed in Table~\ref{tab:datasets} (excluding graph-summary data), obtains the returned graph embeddings, and trains a surrogate encoder via supervised regression. The surrogate is then paired with the frozen text encoder and evaluated for zero-shot node classification on all evaluation datasets (results in Table~\ref{tab:full_domain}).

\noindent \textbf{High-fidelity prediction and strong zero-shot performance.}
Across all evaluation datasets, the attacker models (GPS, GAT, and GCN) consistently achieve strong fidelity, frequently exceeding 80\% and reaching above 90\% in some cases. This indicates that the surrogate encoders closely match the victim GFM’s zero-shot prediction behavior.
In terms of zero-shot classification accuracy, the GAT-based surrogate nearly matches the victim with an average accuracy drop of only 0.15\%. 
The lightweight GCN surrogate also performs competitively, trailing the victim by only 0.66\% on average. 
These results show that contrastive pretraining is \emph{not} required for the attacker to recover the graph-text alignment necessary for zero-shot inference; a single supervised regression objective is sufficient.

In several datasets, the surrogate even surpasses the victim in zero-shot accuracy. 
This may occur because (i) the attacker’s smaller model acts as an implicit regularizer, smoothing decision boundaries and improving generalization, especially when the victim is slightly overfitted to its pretraining data; and (ii) the surrogate learns from the victim’s outputs rather than raw supervision, filtering out noisy or unstable patterns in the victim’s representations and producing cleaner, more stable features.

\noindent \textbf{Security implications.}
These findings show that an adversary can replicate a powerful GFM using only supervised regression and at a small fraction of the original training cost. As GFMs grow more capable and widely deployed, such efficient extraction raises significant intellectual-property and security concerns.

\noindent \textbf{Parameter efficiency in MEA.}
The attacker’s graph encoders contain far fewer parameters than the victim’s encoder, accelerating training and substantially reducing computational cost. This demonstrates that a GFM’s knowledge can be transferred quickly and effectively using only supervised MSE training, enabling extraction into lightweight models that retain high fidelity.

\begin{table*}[t]
  \caption{MEA under query budgets. Comparison of attacker's performance using \textbf{1,000 queries} vs.\ \textbf{500 queries}.}
  \label{tab:query_budget}
  \centering
  \renewcommand{\arraystretch}{0.75}
  \setlength{\tabcolsep}{3pt}
  \small
  \resizebox{\textwidth}{!}{
  \begin{tabular}{c c *{8}{c} *{8}{c}}
    \toprule
    \multirow{3}{*}[-1.8ex]{\textbf{Query Dataset}} &
    \multirow{3}{*}[-1.8ex]{\textbf{Model}} &
    \multicolumn{8}{c}{\textbf{1000 Queries}} &
    \multicolumn{8}{c}{\textbf{500 Queries}} \\
    \cmidrule(lr){3-10} \cmidrule(lr){11-18}
    & & \multicolumn{2}{c}{\textbf{Cora}} & \multicolumn{2}{c}{\textbf{CiteSeer}} & \multicolumn{2}{c}{\textbf{WikiCS}} & \multicolumn{2}{c}{\textbf{Average}} &
        \multicolumn{2}{c}{\textbf{Cora}} & \multicolumn{2}{c}{\textbf{CiteSeer}} & \multicolumn{2}{c}{\textbf{WikiCS}} & \multicolumn{2}{c}{\textbf{Average}} \\
    \cmidrule(lr){3-4}\cmidrule(lr){5-6}\cmidrule(lr){7-8}\cmidrule(lr){9-10}
    \cmidrule(lr){11-12}\cmidrule(lr){13-14}\cmidrule(lr){15-16}\cmidrule(lr){17-18}
    & & \textbf{Acc.} & \textbf{Fid.} & \textbf{Acc.} & \textbf{Fid.} & \textbf{Acc.} & \textbf{Fid.} & \textbf{Acc.} & \textbf{Fid.} &
        \textbf{Acc.} & \textbf{Fid.} & \textbf{Acc.} & \textbf{Fid.} & \textbf{Acc.} & \textbf{Fid.} & \textbf{Acc.} & \textbf{Fid.} \\
    \midrule
    \multirow{3}{*}{\centering \textbf{OGBN-ArXiv}} & GPS
      & 58.55 & \textbf{85.49} & 66.11 & \textbf{89.54} & 66.02 & \textbf{83.48} & 63.56 & \textbf{86.17}
      & 59.41 & 83.89 & 65.79 & \textbf{87.97} & 63.23 & \textbf{79.01} & 62.81 & \textbf{83.62} \\
    & GAT
      & \textbf{67.65} & 82.41 & 67.78 & 88.08 & \textbf{66.33} & 78.18 & \textbf{67.26} & 82.89
      & \textbf{67.65} & 81.55 & \textbf{68.72} & 87.24 & \textbf{65.17} & 76.70 & \textbf{67.18} & 81.83 \\
    & GCN
      & 66.42 & 80.69 & 68.31 & 86.40 & 60.84 & 67.50 & 65.19 & 78.20
      & 66.30 & 79.21 & 66.63 & 84.31 & 59.07 & 63.46 & 64.00 & 75.66 \\
    \hdashline
    \addlinespace[0.8ex]
    \multirow{3}{*}{\centering \textbf{ArXiv-2023}} & GPS
      & 62.36 & 85.24 & 66.53 & 88.39 & 55.88 & 66.05 & 61.59 & 79.89
      & 62.73 & \textbf{84.75} & 65.17 & 84.73 & 52.66 & 59.50 & 60.19 & 76.32 \\
    & GAT
      & 68.76 & 82.04 & 68.62 & 84.00 & 58.79 & 63.80 & 65.39 & 76.61
      & 63.35 & 77.24 & 67.99 & 80.86 & 60.61 & 64.51 & 63.98 & 74.20 \\
    & GCN
      & 68.39 & 78.35 & 67.15 & 82.53 & 56.76 & 56.54 & 64.10 & 72.47
      & 54.00 & 66.42 & 63.60 & 76.15 & 53.06 & 51.55 & 56.89 & 64.71 \\
    \hdashline
    \addlinespace[0.8ex]
    \multirow{3}{*}{\centering \textbf{Pubmed}} & GPS
      & 54.37 & 59.16 & 50.31 & 56.38 & 25.23 & 32.10 & 43.31 & 49.21
      & 51.91 & 55.60 & 47.38 & 55.54 & 32.13 & 39.82 & 43.81 & 50.32 \\
    & GAT
      & 61.01 & 74.78 & 58.26 & 67.89 & 40.73 & 50.10 & 53.33 & 64.26
      & 61.99 & 73.92 & 54.92 & 65.79 & 43.66 & 53.86 & 53.52 & 64.53 \\
    & GCN
      & 33.95 & 44.65 & 38.39 & 46.55 & 21.39 & 27.66 & 31.24 & 39.62
      & 39.61 & 49.82 & 37.03 & 44.46 & 21.73 & 27.83 & 32.79 & 40.70 \\
    \hdashline
    \addlinespace[0.8ex]
    \multirow{3}{*}{\centering \textbf{Mixed Academic}} & GPS
      & 62.61 & 85.36 & 67.05 & 89.44 & 65.59 & 80.38 & 65.08 & 85.06
      & 61.75 & 82.90 & 67.99 & 85.67 & 62.69 & 74.96 & 64.14 & 81.18 \\
    & GAT
      & 65.31 & 80.93 & \textbf{70.08} & 85.25 & 61.78 & 68.64 & 65.72 & 78.28
      & 64.58 & 80.93 & 67.89 & 82.01 & 59.67 & 64.20 & 64.04 & 75.71 \\
    & GCN
      & 66.17 & 79.09 & 67.57 & 84.10 & 55.40 & 60.67 & 63.05 & 74.62
      & 65.68 & 77.61 & 65.59 & 82.11 & 54.69 & 57.08 & 61.98 & 72.27 \\
    \midrule
    \addlinespace[0.8ex]
    \textbf{All}
    & Victim
    & 64.94 & NA
    & 68.83 & NA
    & 70.09 & NA
    & 67.96 & NA
    & 64.94 & NA
    & 68.83 & NA
    & 70.09 & NA
    & 67.96 & NA \\
    \bottomrule
  \end{tabular}
  }
\end{table*}

\begin{table*}[t]
  \caption{MEA under tight query budgets. Comparison of attacker's performance using \textbf{250 queries} and \textbf{100 queries}.}
  \label{tab:query_budget_merged_small}
  \centering
  \renewcommand{\arraystretch}{0.75}
  \setlength{\tabcolsep}{3pt}
  \small
  \resizebox{\textwidth}{!}{
  \begin{tabular}{c c *{8}{c} *{8}{c}}
    \toprule
    \multirow{3}{*}[-1.8ex]{\textbf{Query Dataset}} &
    \multirow{3}{*}[-1.8ex]{\textbf{Model}} &
    \multicolumn{8}{c}{\textbf{250 Queries}} &
    \multicolumn{8}{c}{\textbf{100 Queries}} \\
    \cmidrule(lr){3-10} \cmidrule(lr){11-18}
    & & \multicolumn{2}{c}{\textbf{Cora}} & \multicolumn{2}{c}{\textbf{CiteSeer}} & \multicolumn{2}{c}{\textbf{WikiCS}} & \multicolumn{2}{c}{\textbf{Average}} &
        \multicolumn{2}{c}{\textbf{Cora}} & \multicolumn{2}{c}{\textbf{CiteSeer}} & \multicolumn{2}{c}{\textbf{WikiCS}} & \multicolumn{2}{c}{\textbf{Average}} \\
    \cmidrule(lr){3-4}\cmidrule(lr){5-6}\cmidrule(lr){7-8}\cmidrule(lr){9-10}
    \cmidrule(lr){11-12}\cmidrule(lr){13-14}\cmidrule(lr){15-16}\cmidrule(lr){17-18}
    & & \textbf{Acc.} & \textbf{Fid.} & \textbf{Acc.} & \textbf{Fid.} & \textbf{Acc.} & \textbf{Fid.} & \textbf{Acc.} & \textbf{Fid.} &
        \textbf{Acc.} & \textbf{Fid.} & \textbf{Acc.} & \textbf{Fid.} & \textbf{Acc.} & \textbf{Fid.} & \textbf{Acc.} & \textbf{Fid.} \\
    \midrule
    \multirow{3}{*}{\centering \textbf{OGBN-ArXiv}} & GPS
      & 60.02 & 77.00 & 63.28 & 81.17 & 58.87 & 67.02 & 60.73 & 75.06
      & 51.66 & 65.68 & 60.25 & 73.01 & 40.79 & 43.12 & 50.90 & 60.61 \\
    & GAT
      & \textbf{65.19} & \textbf{80.07} & \textbf{70.61} & \textbf{86.30} & \textbf{61.98} & \textbf{69.30} & \textbf{65.92} & \textbf{78.56}
      & \textbf{63.47} & \textbf{79.34} & \textbf{66.63} & \textbf{80.65} & \textbf{58.42} & \textbf{59.50} & \textbf{62.84} & \textbf{73.16} \\
    & GCN
      & 60.27 & 74.54 & 67.26 & 81.59 & 55.37 & 60.67 & 60.97 & 72.27
      & 49.20 & 58.92 & 60.15 & 73.01 & 54.37 & 53.52 & 54.57 & 61.82 \\
    \hdashline
    \addlinespace[0.8ex]
    \multirow{3}{*}{\centering \textbf{ArXiv-2023}} & GPS
      & 63.84 & 78.47 & 63.18 & 78.77 & 48.65 & 53.86 & 58.55 & 70.37
      & 43.42 & 58.79 & 54.18 & 63.49 & 25.55 & 28.17 & 41.05 & 50.15 \\
    & GAT
      & 55.84 & 66.91 & 67.47 & 77.72 & 51.18 & 55.63 & 58.16 & 66.75
      & 46.00 & 57.93 & 63.91 & 76.15 & 39.25 & 40.81 & 49.72 & 58.30 \\
    & GCN
      & 52.77 & 60.27 & 61.72 & 73.12 & 43.29 & 44.97 & 52.59 & 59.45
      & 39.11 & 47.23 & 49.48 & 60.04 & 23.75 & 22.13 & 37.45 & 43.13 \\
    \hdashline
    \addlinespace[0.8ex]
    \multirow{3}{*}{\centering \textbf{Pubmed}} & GPS
      & 39.73 & 43.42 & 46.13 & 52.09 & 23.90 & 29.36 & 36.59 & 41.63
      & 31.61 & 31.49 & 27.30 & 32.95 & 20.02 & 22.96 & 26.31 & 29.13 \\
    & GAT
      & 57.69 & 67.40 & 53.35 & 64.64 & 46.31 & 56.37 & 52.45 & 62.80
      & 53.87 & 65.31 & 43.93 & 52.20 & 29.19 & 35.06 & 42.33 & 50.86 \\
    & GCN
      & 37.88 & 40.47 & 41.95 & 50.00 & 28.40 & 34.52 & 36.08 & 41.66
      & 42.07 & 45.51 & 31.28 & 36.51 & 21.65 & 24.18 & 31.66 & 35.40 \\
    \hdashline
    \addlinespace[0.8ex]
    \multirow{3}{*}{\centering \textbf{Mixed Academic}} & GPS
      & 56.70 & 76.14 & 64.02 & 80.75 & 54.20 & 60.75 & 58.31 & 72.55
      & 51.41 & 58.55 & 44.35 & 48.54 & 37.57 & 36.49 & 44.44 & 47.86 \\
    & GAT
      & 62.85 & 78.84 & 67.99 & 84.10 & 56.14 & 60.87 & 62.33 & 74.60
      & 60.76 & 75.40 & 64.33 & 69.25 & 48.08 & 48.31 & 57.72 & 64.32 \\
    & GCN
      & 62.61 & 74.78 & 64.12 & 78.45 & 54.23 & 55.71 & 60.32 & 69.65
      & 56.83 & 63.96 & 46.23 & 50.84 & 39.87 & 36.26 & 47.65 & 50.35 \\
    \bottomrule
  \end{tabular}
  }
\end{table*}

\subsubsection{Extracting Domain-Specific Knowledge from a GFM}

The victim GFM is pretrained on graphs from multiple domains (academic, e-commerce, and social). The attacker’s goal is to extract domain-specific knowledge from the GFM into a smaller graph encoder. To achieve this, the attacker queries the victim’s graph encoder using domain-specific training graphs and uses the returned embeddings as supervision to train its encoder. Evaluation is then conducted on unseen graphs from the same domain (results shown in Table~\ref{tab:domain_specific}).
To extract academic-domain knowledge, the attacker constructs queries using the open-source datasets \emph{OGBN-ArXiv}, \emph{ArXiv-2023}, and \emph{PubMed}, and trains its encoder on the corresponding victim embeddings. Zero-shot performance is then evaluated on \emph{Cora}, \emph{CiteSeer}, and \emph{WikiCS}. 
Analogously, e-commerce knowledge is extracted by training on \emph{OGBN-Products} and evaluating on \emph{Ele-Photo}, \emph{Ele-Computers}, and \emph{Books-History}.  
For the social domain, the attacker uses \emph{Reddit} to query the victim and tests on \emph{Instagram}.

\noindent \textbf{Architecture mismatch.} 
To emulate a realistic adversary, we assume the attacker does not know the victim’s architecture. Accordingly, we replace the victim's GPS layers with alternative GNN layers (GAT and GCN) in the surrogate and perform the extraction under this architectural mismatch. We then evaluate zero-shot transfer both within-domain and across domains.
The results show that surrogates achieve high fidelity and near-victim zero-shot accuracy on graphs from the target domain—even without architectural knowledge. 

More interestingly, we find that a GNN-based attacker, which is typically limited by local message passing and over-smoothing, still attains competitive zero-shot generalization to new graphs. This is surprising because GNNs are generally considered less expressive than graph transformers, which leverage attention and weaker inductive bias to capture long-range dependencies and serve as more expressive, pretrainable architectures. Our results indicate that, despite their simplicity, GNNs can successfully inherit and transfer domain-level graph-text knowledge from a GFM, achieving strong performance on unseen graphs without relying on the victim’s architectural details.

\noindent \textbf{Knowledge transfer.} 
We also evaluate the attacker on graphs from domains \emph{different} from the one used for training. Table~\ref{tab:domain_specific} shows a clear pattern: surrogates perform best on graphs from the same domain as their query set, achieving substantially higher zero-shot accuracy and fidelity compared to out-of-domain tests. This confirms that querying the victim with domain-specific graphs allows the attacker to selectively extract that domain's knowledge and transfer it effectively to new graphs.

\subsubsection{MEA under Query Budget}
\begin{figure}[t]
   \begin{center}
   \includegraphics[width=1.0\linewidth]{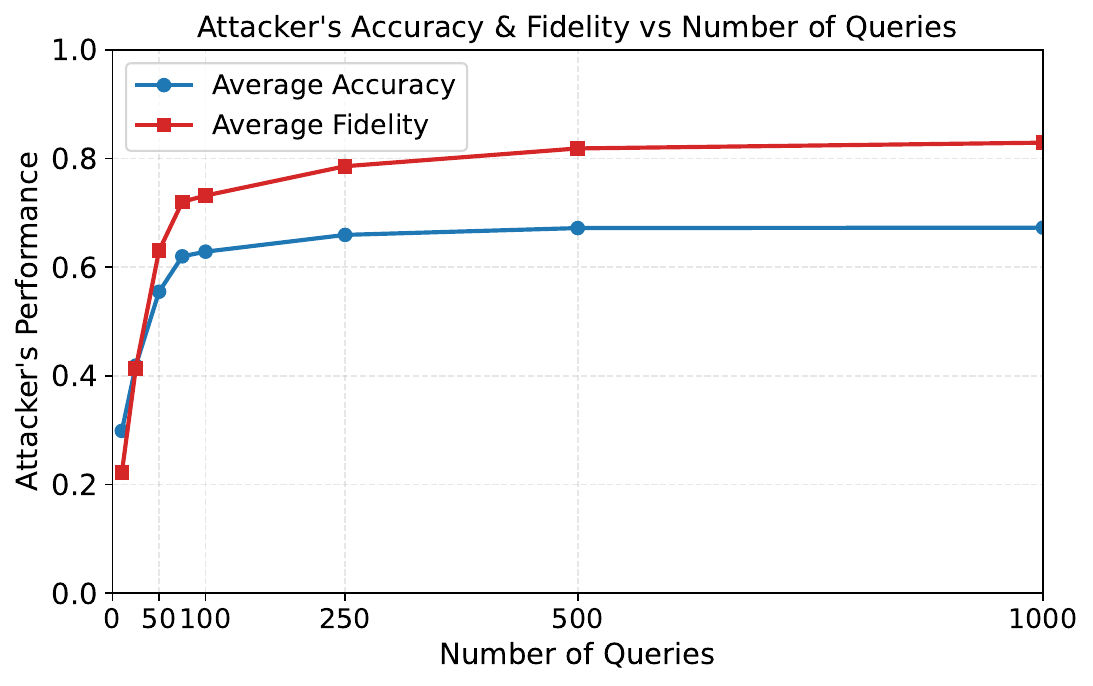} %
    \end{center}
    \vspace{-0.15in}
 \caption{Average attacker accuracy and fidelity vs.\ number of queries (GAT attacker).}
\label{fig:queries}
\vspace{-0.1in}
\end{figure}

We follow the setup in Section~\ref{subsec:exp_setup}, where queries to the victim model are limited or costly, so the attacker must maximize the utility of a constrained query budget. Table~\ref{tab:query_budget}, Table~\ref{tab:query_budget_merged_small} and Fig. \ref{fig:queries} report results for different query \emph{sources} and different \emph{numbers} of queries. The results reveal a clear dependency on both factors. 
First, the choice of query source matters: using \textit{ogbn-ArXiv} yields the strongest surrogates, with the GAT-based attacker reaching an average accuracy of $0.6726$ and fidelity of $0.8289$—closely matching the victim’s zero-shot performance. In contrast, querying only \textit{PubMed} leads to large drops in accuracy and fidelity across attacker models \emph{despite} PubMed being part of the victim’s pretraining corpus. This suggests that extraction quality depends more on the similarity between the attacker’s query distribution and the \emph{evaluation} graphs. When the query source (e.g., PubMed) is semantically or structurally farther from the target evaluation datasets (e.g., Cora, CiteSeer, WikiCS), knowledge transfer becomes harder and the surrogate underperforms.

Second, even when the attacker reduces the query budget from $1000$ to $250$ or $100$ queries, the attack remains effective. Although performance declines as the budget shrinks, the GAT attacker trained on \textit{OGBN-ArXiv} still preserves high victim agreement, achieving average fidelities above $0.78$ with $250$ queries and above $0.73$ with only $100$ queries. This shows that strong surrogates can be obtained even when the attacker can afford only a small number of expensive model queries.
Third, the mixed-dataset strategy performs competitively—often outperforming \textit{ArXiv-2023}-only and substantially surpassing \textit{PubMed}-only—yet still lags behind \textit{OGBN-ArXiv}-only queries. Increasing the number of training epochs under smaller budgets (500, 250, and 100 queries) improves stability and accuracy, but cannot fully bridge the gap when the supervision obtained from queries is less relevant to the evaluation datasets.

\subsubsection{Graph-Specific Extraction Attack}
\begin{figure}[t]
   \begin{center}
   \includegraphics[width=1.0\linewidth]{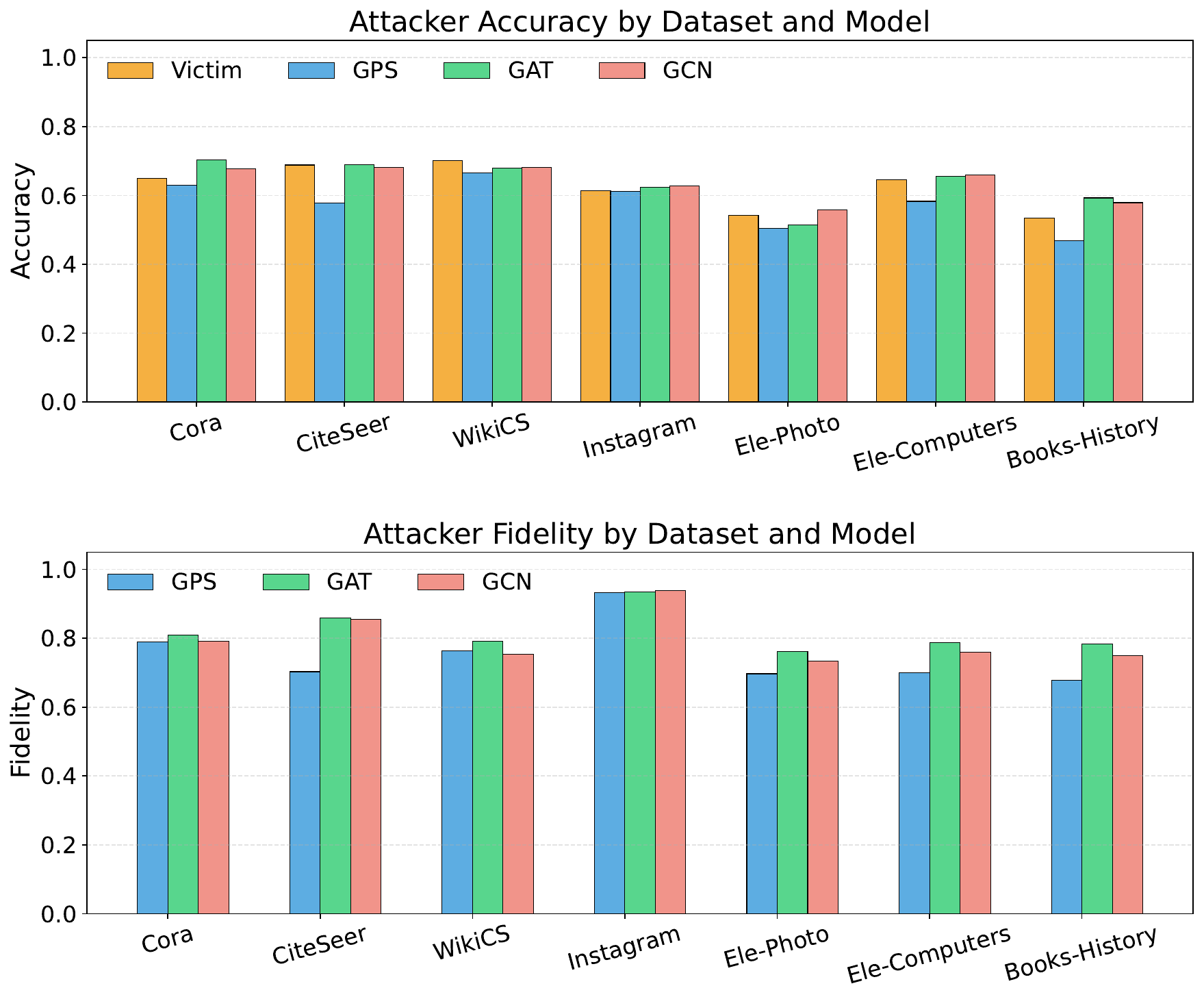}
    \end{center}
 \caption{Performance of graph-specific extraction attacks.}
\label{fig:graph_specific}
\end{figure}

Since a pretrained GFM can perform zero-shot inference on graphs that were not used during pretraining, the attacker can directly use a target graph to construct queries for the victim GFM and train a surrogate model. This setting resembles prior MEA work on GNNs \cite{wu2022model, shen2022model, wang2025cega, zhuang2024unveiling} where the attacker aims to train a graph-specific surrogate for a particular target graph. A key difference is that, because GFMs generalize across graphs, an attacker can extract different surrogates tailored to different target graphs. For each evaluation dataset (Table~\ref{tab:datasets}), we sample a subset of nodes as queries, obtain their embeddings from the victim model, train the attacker model on these embeddings, and then evaluate the surrogate on the held-out test nodes from the same graph. This procedure matches the semi-supervised setup commonly used in GNN benchmarks. The results are shown in Fig. \ref{fig:graph_specific}. 

We observe that when the attacker does not need to extract full domain-level knowledge, the extraction task becomes even easier. By querying the victim GFM with only a small subset of nodes from the target graph, the attacker can train a surrogate model whose performance closely matches that of the victim on the same graph. In several datasets, the surrogate even outperforms the victim, despite using far fewer parameters and significantly less training time. These findings indicate that graph-specific extraction requires minimal effort but still yields high-fidelity surrogates, underscoring that large-scale graph models are at serious risk even in the simplest attack scenario.

\begin{table*}[t]
  \caption{MEA with synthetic graphs. The attacker can successfully replicate the victim GFM even when only a small subset of node attributes is observable, by synthesizing missing features to form queryable subgraphs.}
  \label{tab:synthetic_graph}
  \centering
  \renewcommand{\arraystretch}{1.1}
  \setlength{\tabcolsep}{4pt}
  \resizebox{\textwidth}{!}{
  \begin{tabular}{c|*{7}{cc}}
    \hline
    \multirow{2}{*}{\textbf{Model}} &
    \multicolumn{2}{c}{\textbf{Cora}} &
    \multicolumn{2}{c}{\textbf{CiteSeer}} &
    \multicolumn{2}{c}{\textbf{WikiCS}} &
    \multicolumn{2}{c}{\textbf{Instagram}} &
    \multicolumn{2}{c}{\textbf{Ele-Photo}} &
    \multicolumn{2}{c}{\textbf{Ele-Computers}} &
    \multicolumn{2}{c}{\textbf{Books-History}} \\
    \cline{2-15}
    & \textbf{Acc.} & \textbf{Fid.}
    & \textbf{Acc.} & \textbf{Fid.}
    & \textbf{Acc.} & \textbf{Fid.}
    & \textbf{Acc.} & \textbf{Fid.}
    & \textbf{Acc.} & \textbf{Fid.}
    & \textbf{Acc.} & \textbf{Fid.}
    & \textbf{Acc.} & \textbf{Fid.} \\
    \hline
    GPS
    & 62.73 & \textbf{86.22}
    & \textbf{68.10} & \textbf{95.29}
    & \textbf{69.21} & \textbf{91.88}
    & \textbf{62.40} & \textbf{94.18}
    & 51.60 & \textbf{86.09}
    & 61.97 & \textbf{86.63}
    & \textbf{53.45} & \textbf{92.71} \\
    GAT
    & 67.04 & 82.29
    & 67.78 & 90.79
    & 67.87 & 82.65
    & 60.55 & 88.62
    & 55.75 & 78.97
    & 64.84 & 81.84
    & 47.74 & 83.98 \\
    GCN
    & \textbf{67.28} & 80.69
    & 66.84 & 88.81
    & 67.27 & 79.64
    & 61.20 & 88.15
    & \textbf{57.32} & 74.78
    & \textbf{65.09} & 78.94
    & 53.29 & 84.42 \\
    \hline
    Victim
    & 64.94 & NA
    & 68.83 & NA
    & 70.09 & NA
    & 61.29 & NA
    & 54.13 & NA
    & 64.46 & NA
    & 53.32 & NA \\
    \hline
  \end{tabular}
  }
\end{table*}

\begin{table*}[t]
  \caption{Even without access to the victim’s pretraining datasets, the attacker can query the victim model using public graphs and build a surrogate with high fidelity and strong zero-shot performance. All values are percentages (\%).
  }
  \label{tab:extra_dataset}
  \centering
  \renewcommand{\arraystretch}{1.1}
  \setlength{\tabcolsep}{4pt}
  \resizebox{\textwidth}{!}{
  \begin{tabular}{c|*{7}{cc}}
    \hline
    \multirow{2}{*}{\textbf{Model}} &
    \multicolumn{2}{c}{\textbf{Cora}} &
    \multicolumn{2}{c}{\textbf{CiteSeer}} &
    \multicolumn{2}{c}{\textbf{WikiCS}} &
    \multicolumn{2}{c}{\textbf{Instagram}} &
    \multicolumn{2}{c}{\textbf{Ele-Photo}} &
    \multicolumn{2}{c}{\textbf{Ele-Computers}} &
    \multicolumn{2}{c}{\textbf{Books-History}} \\
    \cline{2-15}
    & \textbf{Acc.} & \textbf{Fid.}
    & \textbf{Acc.} & \textbf{Fid.}
    & \textbf{Acc.} & \textbf{Fid.}
    & \textbf{Acc.} & \textbf{Fid.}
    & \textbf{Acc.} & \textbf{Fid.}
    & \textbf{Acc.} & \textbf{Fid.}
    & \textbf{Acc.} & \textbf{Fid.} \\
    \hline
    GPS
    & \textbf{67.16} & \textbf{85.49}
    & 66.21 & \textbf{88.81}
    & \textbf{67.25} & \textbf{89.26}
    & \textbf{58.26} & \textbf{86.33}
    & 47.38 & \textbf{70.56}
    & 52.89 & 69.31
    & 53.10 & \textbf{89.60} \\
    GAT
    & 62.85 & 80.44
    & \textbf{66.63} & 86.09
    & 65.11 & 82.03
    & 54.12 & 68.61
    & \textbf{52.66} & 70.23
    & \textbf{64.81} & \textbf{79.45}
    & 56.49 & 83.37 \\
    GCN
    & 47.72 & 59.29
    & 61.92 & 78.56
    & 61.07 & 72.00
    & 48.97 & 48.59
    & 50.92 & 66.40
    & 58.44 & 71.35
    & \textbf{56.95} & 78.08 \\
    \hline
    Victim
    & 64.94 & NA
    & 68.83 & NA
    & 70.09 & NA
    & 61.29 & NA
    & 54.13 & NA
    & 64.46 & NA
    & 53.32 & NA \\
    \hline
  \end{tabular}
  }
  \vspace{-0.03in}
\end{table*}

\subsubsection{MEA with Synthetic Graphs}
Table~\ref{tab:synthetic_graph} summarizes results when the attacker has access only to a small subset of node attributes and their local neighborhoods in each training graph. Overall, these results show that model extraction remains highly effective even with only partial visibility of the training graphs. All surrogate models achieve reasonably strong fidelity and zero-shot performance, demonstrating that homophily-based synthetic queries still provide meaningful supervision despite incomplete node attributes. Among them, the GPS surrogate attains the highest fidelity across all datasets, while lightweight GNNs such as GCN and GAT remain competitive in accuracy and, on several datasets (\textit{Cora}, \textit{Ele-Photo}, \textit{Ele-Computers}), even outperform GPS in zero-shot inference. This suggests that smooth message passing provides a useful inductive bias when synthesized features approximate local neighborhood averages. Moreover, all surrogates preserve strong zero-shot transfer on unseen graphs, indicating that alignment with the frozen text encoder remains intact despite partial visibility and synthesized attributes.

\noindent \textbf{Takeaways and limitations.}
Homophily-based synthesis provides a simple and effective mechanism to unlock MEA under partial node access: attackers do not need proprietary graph-text pairs nor full attribute visibility to approximate the victim model. GPS excels at reproducing the victim’s decisions (high \emph{Fidelity}), while compact GCN/GAT surrogates often match the victim’s \emph{Accuracy} with far fewer parameters. One limitation is that the synthesis rule assumes moderate homophily; graphs with weak or heterophilous structure may require more expressive imputers or learned synthesis methods, which we leave for future work.

\subsubsection{MEA without Using Training Datasets of the Victim Model}
In Attack-6, the attacker does not know which datasets the victim GFM was pretrained on and instead queries the model using other publicly available graphs, resulting in a training data discrepancy between the attacker’s queries and the victim’s original training data. As shown in Table~\ref{tab:extra_dataset}, MEA remains highly effective: the resulting surrogates achieve strong zero-shot accuracy and high fidelity on unseen graphs, despite never querying the victim with its true pretraining datasets.
The GPS-based surrogate obtains the highest fidelity across most evaluation graphs and maintains zero-shot accuracy that is close to the victim model. On \textit{Cora}, it even surpasses the victim in accuracy. Lightweight GAT and GCN surrogates are also effective: despite querying mismatched datasets, both models achieve competitive zero-shot accuracy across domains. Notably, in this setting the attacker queries only 3 public datasets to train the surrogate, yet reaches performance comparable to the victim model pretrained on 5 large-scale datasets.

\noindent \textbf{Security implications.}
These results show that MEA does not require access to the victim’s actual pretraining graphs. As long as the attacker can collect any reasonably related attributed graphs from the same broader domain, supervised regression on the returned embeddings is enough to reconstruct a high-fidelity surrogate. In practice, this means that even if a model owner hides or restricts access to the datasets used for pretraining, an attacker can still extract the victim’s knowledge using publicly available data sources.
This finding further expands the real-world risk surface of GFMs: restricting access to original pretraining graphs is not sufficient to prevent model extraction. Deployment-level defenses must instead address the security of embedding APIs themselves, without assuming that hiding training data protects the model.

\begin{table}[t]
  \caption{Attacker model size and computational cost.}
  \label{tab:cost_analysis}
  \resizebox{\columnwidth}{!}{
  \begin{tabular}{c|c c}
    \hline
    \textbf{Model} & \textbf{Trainable Parameter} & \textbf{GPU Training Time} \\
    \hline
     Victim Model & 128,007,618 & 56.00 (h) \\
     GPS Attacker & 11,102,400 & 0.0833 (h) \\
     GAT Attacker & 2,931,456 & 0.0389 (h) \\
     GCN Attacker & 952,064 & 0.02917 (h) \\
    \hline
  \end{tabular}
  }
\end{table}

\subsection{Attacker Model Size and Computational Cost}
To understand the computational feasibility of MEA against large GFMs, we compare the parameter counts and training time of the victim and attacker models. Table~\ref{tab:cost_analysis} reports the number of trainable parameters and total GPU training hours required for MEA under the Attack-1 setting.

We observe a striking difference in resource requirements. The victim GFM contains 128M trainable parameters and requires 56 GPU hours to train, whereas the GPS-based attacker has only 11M parameters and trains in about 5 minutes on a single GPU. The GAT and GCN attackers are even smaller (2.9M and 0.95M parameters) and complete training in under 3 minutes and under 2 minutes, respectively. In total, the GPS attacker requires only 0.15\% of the victim’s training time, the GAT attacker 0.07\%, and the GCN attacker just 0.05\%.
Despite this enormous gap in model size and training cost, the attacker models maintain high fidelity and accurate zero-shot inference performance across all evaluation datasets. These results reveal a concerning gap:
\begin{itemize}[leftmargin=*]
    \item Training the victim GFM requires massive computation and large-scale data.
    \item Yet an attacker can cheaply replicate its function using only supervised regression on queried embeddings.
\end{itemize}

In short, GFMs are expensive to build but inexpensive to steal. This amplifies the practical risk of model extraction and motivates the development of stronger, deployment-aware defenses for large-scale multimodal graph learning systems.

\section{Discussion}
\subsection{Countermeasures}
While GFMs enable powerful zero-shot generalization, their deployment exposes rich embedding outputs that can be exploited for model extraction. Because attackers can approximate a victim's graph encoder purely from returned embeddings, effective defenses must reduce information leakage while preserving practical utility for legitimate users.

\noindent
\textbf{(1) Output Perturbation and Truncation.}
A direct defense strategy is to limit the precision or volume of model outputs. 
Instead of returning full graph embeddings, a service may:
(i) provide only similarity scores from the two encoders,
(ii) truncate or project embeddings into a lower-dimensional space, or
(iii) add small stochastic perturbations (e.g., Gaussian or Laplacian noise).
These mechanisms reduce the fidelity with which an attacker can approximate the victim encoder.
Noise injection aligns with differential privacy principles, though excessive perturbation may degrade zero-shot accuracy, requiring a careful privacy-utility trade-off.

\noindent
\textbf{(2) Query Monitoring and Rate Limiting.}
MEA methods typically require a large number of diverse queries.
Thus, defensive systems may:
(i) enforce strict rate limits per user or session,
(ii) detect abnormal or automated querying patterns, or
(iii) limit repeated queries to structurally similar subgraphs.
Restricting query volume and diversity constrains the attacker’s ability to approximate the victim mapping, while preserving usability for standard inference workloads.

\noindent
\textbf{(3) Output Watermarking.}
Embedding-space watermarking introduces imperceptible but structured perturbations into returned outputs.
Such watermarks can embed a cryptographic signature so that, if a surrogate later reproduces \nolinebreak the signature, the model owner can prove unauthorized duplication.
Watermarking provides post-attack attribution rather than prevention, and can be combined with the above defenses.

\subsection{Future Directions} 
While this work provides a systematic study of MEA against GFMs, several promising directions remain for extending this line of research. These directions involve substantially different technical challenges or require assumptions and infrastructures beyond the threat model considered here, and are therefore left for future investigation.
(1) studying MEA when the GFM exposes only similarity scores rather than graph embeddings, which requires new learning objectives that operate on low-dimensional, partial, or noisy feedback; (2) extending MEA to other multimodal foundation models (e.g., time-series–language, point-cloud–language) to understand which architectural or training designs offer greater resilience; (3) improving query efficiency through active sampling or subgraph summarization to reduce API calls; and (4) developing feedback-driven querying, where future queries are chosen based on current surrogate errors or disagreement with the victim, enabling an active extraction loop that maintains high fidelity under tight budgets. Additional details about future directions are in Appendix~\ref{appen:future}.
\section{Related Work}
\subsection{Model Extraction Attack against GNNs}
Earlier studies on MEAs in graph learning have primarily focused on conventional GNNs operating in \textbf{transductive} settings~\cite{wu2022model, defazio2019adversarial}. These methods typically assume that the adversary possesses partial or complete knowledge of the victim’s training environment, including node attributes, global topology, or subgraph connectivity. For example,~\cite{wu2022model} proposed an MEA framework that reconstructs a target GNN by synthesizing realistic query nodes and exploiting both their returned predictions and structural priors from the underlying graph. Their work systematically defined seven threat models reflecting different levels of attacker knowledge and proposed tailored extraction strategies for each. More recently, CEGA~\cite{wang2025cega} examined GNN extraction in MLaaS settings and demonstrated that adaptive node querying can reconstruct high-fidelity surrogates under limited access, revealing both critical security vulnerabilities and potential for label-efficient graph learning.

A parallel line of research has explored MEAs in \textbf{inductive} graph learning, where the attacker must operate on unseen nodes or entirely new graphs. The first systematic effort in this direction,~\cite{shen2022model}, introduced a comprehensive threat model for inductive GNNs and proposed six attack variants based on different levels of adversarial knowledge and model feedback. Building on this, \textsc{StealGNN}~\cite{zhuang2024unveiling} relaxed assumptions on data availability by removing the need for node features drawn from the same distribution as the victim’s training data. Their framework introduced the first data-free MEA for GNNs, expanding the scope of feasible attacks under more constrained and realistic conditions.
Despite these advancements, prior approaches remain limited in both scale and theoretical depth. They focus on small-scale GNNs and do not address the security implications of modern large-scale graph models. This paper takes the first step toward investigating MEAs on GFMs, unveiling a new class of threats that enable domain-level knowledge theft and zero-shot generalization across graphs.

\subsection{Knowledge Distillation in Multimodal Models}
Knowledge distillation (KD) transfers representational and functional knowledge from a large, well-trained teacher model to a smaller student model, serving as a cornerstone technique for model compression and performance enhancement. In multimodal learning, most KD research focuses on distilling vision-language models (VLMs) \cite{li2025ammkd, liu2025mllm4ts, li2025catp} and has become increasingly important for replicating cross-modal alignment while reducing model size. CLIP-KD \cite{yang2024clip} provides a systematic empirical study of distillation objectives for VLMs, comparing multiple combinations of feature-level, contrastive, and relational losses. Building on this direction, TinyCLIP \cite{wu2023tinyclip} introduces a progressive distillation framework that leverages weight inheritance and multi-stage training to produce compact student models—achieving up to a fourfold reduction in size compared to the original ViT-B/32 CLIP teacher while maintaining competitive multimodal performance.
\textbf{In the MEA setting, the attacker typically operates under far more restrictive conditions than in standard KD.} Nevertheless, we demonstrate that a surrogate graph encoder can be trained to imitate a GFM using only a supervised distillation loss, effectively transferring domain-level graph-text knowledge without access to the victim’s parameters or training data.


\section{Conclusion}
This paper presents the first systematic study of model extraction attacks against GFMs. We formalize realistic attacker settings, develop a six-way taxonomy that spans domain-level and graph-specific objectives, architectural mismatch, query budgets, partial node access, and training data discrepancies, and propose an effective supervised embedding-regression framework. Despite having no access to proprietary graph-text pretraining data, our approach trains a surrogate graph encoder that remains well aligned with the public text encoder and preserves the victim model’s zero-shot inference on entirely unseen graphs. Theory and experiments jointly show that matching the victim’s graph embeddings on representative queries is sufficient to retain zero-shot decision behavior, revealing a significantly expanded extraction risk surface for large multimodal graph systems.

\bibliographystyle{IEEEtran}
\bibliography{reference}

\appendices
\newpage

\begin{figure*}[!t]
   \begin{center}
      \includegraphics[width=0.9\linewidth]{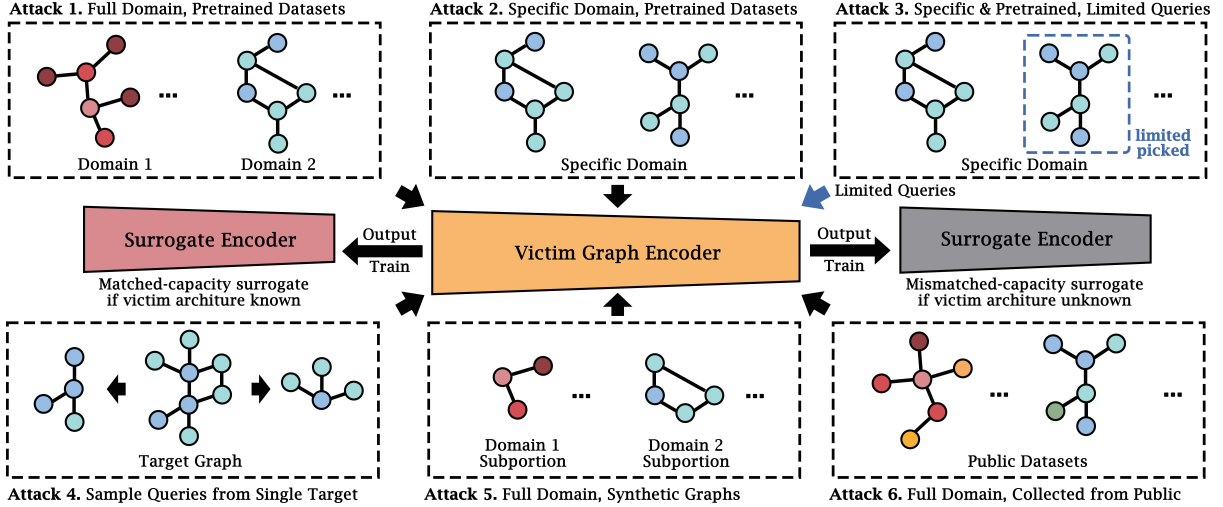} %
    \end{center}
 \caption{Six practical model extraction attack scenarios on GFMs.}
\label{fig:taxonomy}
\end{figure*}

\section{Datasets}
\noindent In this section, we present a comprehensive overview of the datasets utilized in this paper.

\subsection{Pretraining and Evaluation Datasets}
\noindent We use the pretraining and evaluation datasets released with GraphCLIP \cite{zhu2025graphclip}. The original resources are accessible at \url{https://github.com/ZhuYun97/GraphCLIP}.

\noindent\textbf{OGBN-ArXiv.} A directed citation graph of computer science arXiv papers indexed by the Microsoft Academic Graph \cite{wang2020microsoft}. Each node is a paper and each directed edge indicates a citation. The objective is to predict one of 40 subject areas such as cs.AI, cs.LG, and cs.OS, with labels provided by authors and arXiv moderators.

\noindent\textbf{ArXiv-2023.} A directed graph capturing the citation network among computer science arXiv papers published in 2023 or later. Nodes represent papers and directed edges denote citations. The task is to classify each paper into one of 40 subject areas (e.g., cs.AI, cs.LG, cs.OS), using labels assigned by the authors and arXiv moderators.

\noindent\textbf{PubMed.} A three-class dataset focused on diabetes research: (1) experimental studies on disease mechanisms and therapies, (2) Type 1 diabetes work emphasizing autoimmune processes and treatments, and (3) Type 2 diabetes studies centered on insulin resistance and management.

\noindent\textbf{OGBN-Products.} A large-scale product co-purchase network with roughly 2 million nodes and 61 million edges. Nodes correspond to Amazon products and edges reflect co-purchase relationships. The prediction task is to assign each product to one of 47 top-level categories.

\noindent\textbf{Reddit.} A social graph where nodes correspond to users, node features encode the content of users’ previously posted subreddits, and edges indicate reply interactions between users.

\noindent\textbf{Cora.} Contains 2,708 scientific publications categorized into seven topics—case-based, genetic algorithms, neural networks, probabilistic methods, reinforcement learning, rule learning, and theory. Papers form a citation network with 5,429 edges, where each paper cites or is cited by at least one other paper.

\noindent\textbf{CiteSeer.} Comprises 3,186 publications categorized into six areas: Agents, Machine Learning, Information Retrieval, Databases, Human-Computer Interaction, and Artificial Intelligence. The goal is to classify each paper using its title and abstract.

\noindent\textbf{WikiCS.} A Wikipedia-derived benchmark for graph neural networks with 10 computer science branches as classes. Node features are extracted from the corresponding article texts, and the graph exhibits high connectivity.

\noindent\textbf{Instagram.} A social network in which nodes denote users and edges represent following relationships. The task is to classify users as commercial or regular.

\noindent\textbf{Ele-Photo.} Derived from Amazon-Electronics corpus. Nodes represent electronic products and edges capture frequent co-purchases or co-views. Each node carries a text attribute—the highest-voted user review (or a randomly selected review if none is highly voted). Labels follow a three-level electronics taxonomy, and the task is to classify products into 12 categories.

\noindent\textbf{Ele-Computers.} Also derived from Amazon-Electronics corpus with the same graph construction and text attributes as Ele-Photo. The classification task assigns products to 10 categories under a three-level electronics taxonomy.

\noindent\textbf{Books-History.} Extracted from Amazon-Books corpus and restricted to items labeled “History.” Nodes are books; edges encode frequent co-purchases or co-views. Each node uses the book’s title and description as text attributes. Labels follow a three-level taxonomy, and the task is to classify books into 12 categories.

\subsection{Extra Query Datasets}
\noindent For three extra datasets, we adopt the processed versions provided by TSGFM \cite{chen2024text}, available at \url{https://github.com/CurryTang/TSGFM}.

\noindent\textbf{DBLP.} A bibliographic graph of computer science publications. It is originally proposed by \cite{10.1007/978-3-642-15880-3_42}. TSGFM \cite{chen2024text} formulates it as a four-class classification benchmark by constructing a paper–paper graph from co-authorship. 

\noindent\textbf{Books-Children.} Derived from the Amazon-Books corpus, this graph comprises items tagged with the second-level category “Children.” Nodes represent books, and edges link pairs of books that are frequently co-purchased or co-viewed. Labels follow a three-level taxonomy assigned to each book. The node text features are the book’s title and description. The predictive task is a 24-way classification of books into the specified categories.

\noindent\textbf{Sports-Fitness.} A graph derived from the Amazon-Sports corpus, comprising items tagged with the second-level category “Fitness.” Nodes represent fitness-related products, and edges connect pairs of items that are frequently co-purchased or co-viewed. Each node is annotated with a three-level product taxonomy. The predictive task is a 13-class item classification.

\section{Attack Taxonomy}
\label{appen:taxonomy}
We illustrate the six practical MEA scenarios for GFMs in Fig.~\ref{fig:taxonomy}.

\section{Proof of Lemma 1}
\label{appen:proof}

For an unseen test node $v$, define
\[
a = g_{\text{attack}}(v), \qquad b = g_{\text{victim}}(v),
\]
and let
\[
c_{\text{attack}}(v) = \arg\max_{k} \langle a, Z_k \rangle,
\qquad
c_{\text{victim}}(v) = \arg\max_{k} \langle b, Z_k \rangle.
\]
where $Z_k \in \mathbb{R}^d$ denotes the (unit-normalized) text embedding of the $k$-th zero-shot class description.
Consider the margin
\begin{equation}
\Delta = \langle a, Z_{c_{\text{attack}}(v)} \rangle - \langle a, Z_{c_{\text{victim}}(v)} \rangle .
\label{eq:delta_def}
\end{equation}
Adding and subtracting $b$ yields
\begin{equation}
\Delta
=
\langle a-b,\; Z_{c_{\text{attack}}(v)} - Z_{c_{\text{victim}}(v)} \rangle
+
\big( \langle b, Z_{c_{\text{attack}}(v)} \rangle - \langle b, Z_{c_{\text{victim}}(v)} \rangle \big).
\label{eq:decomp}
\end{equation}
Since $c_{\text{victim}}(v)$ maximizes $\langle b, Z_k\rangle$, the second term is non-positive.
Using Cauchy–Schwarz,
\begin{equation}
\Delta
\le
\|a-b\|_2 \cdot \|Z_{c_{\text{attack}}(v)} - Z_{c_{\text{victim}}(v)}\|_2 .
\label{eq:cs}
\end{equation}

Let $\tilde v$ be the node from the training set guaranteed by Assumption~\eqref{eq:assump1} and denote
\[
\tilde a = g_{\text{attack}}(\tilde v),\qquad \tilde b = g_{\text{victim}}(\tilde v).
\]
The triangle inequality and Assumptions~\eqref{eq:assump1}–\eqref{eq:assump2} imply
\begin{equation}
\|a-b\|_2
\le
\|a-\tilde a\|_2 +
\|\tilde a-\tilde b\|_2 +
\|\tilde b-b\|_2
< 3\varepsilon .
\label{eq:triangle}
\end{equation}
Substituting \eqref{eq:triangle} into \eqref{eq:cs} gives
\begin{equation}
\Delta < 3\,\|Z_{c_{\text{attack}}(v)} - Z_{c_{\text{victim}}(v)}\|_2\,\varepsilon .
\label{eq:tight}
\end{equation}
For unit-normalized label embeddings, $\|Z_{c_{\text{attack}}(v)} - Z_{c_{\text{victim}}(v)}\|_2 \le 2$, and therefore
\begin{equation}
\langle g_{\text{attack}}(v), Z_{c_{\text{attack}}(v)} \rangle
-
\langle g_{\text{attack}}(v), Z_{c_{\text{victim}}(v)} \rangle
< 6\varepsilon ,
\end{equation}
which matches the stated bound.
\hfill$\square$

\section{Detailed Future Directions}
\label{appen:future}
\begin{itemize}[leftmargin=*]
    \item \textbf{Similarity scores only.} Investigate MEA when the GFM provider exposes only similarity scores with respect to a set of textual prototypes or label prompts, rather than graph embeddings. This setting requires developing learning objectives that operate purely on score vectors and establishing identifiability guarantees when feedback is partial, low-dimensional, or noisy.
    \item \textbf{Beyond graph-text GFMs.} Extend the framework to other multimodal foundation models that pair a modality-specific encoder with a public text encoder, such as time series-language and point cloud-language systems. A unified treatment may clarify which architectural and training choices are inherently more resilient.
    \item \textbf{Query efficiency.} Develop more query-efficient strategies that construct representative queries with strong coverage of the victim’s embedding space. Promising directions include active sampling, subgraph summarization, and coverage-aware batching to reduce the number of API calls required to train a high-fidelity surrogate.
    \item \textbf{Feedback-driven querying.} Incorporate iterative feedback during query selection. For example, adaptively choose the next batch of queries based on current surrogate errors, uncertainty, or disagreement with the victim, enabling an active MEA loop that improves fidelity under budgets.
\end{itemize}

We hope these directions inspire further research on GFM security and defenses, including output policies, monitoring, and auditing, so that practical deployments remain useful while minimizing extraction risk.

\section*{LLM Usage Considerations}
In accordance with the conference policy on the use of LLMs, we describe and justify any LLM involvement in preparing this paper. 

\noindent \textbf{Originality.} LLMs were used for editorial purposes only, such as polishing grammar, improving clarity, and rephrasing text for readability. All scientific content, ideas, algorithms, theoretical results, figures, experiments, and analysis were developed independently by the authors. All LLM-generated text was reviewed, verified, and edited to ensure accuracy, originality, and proper attribution. The literature review, problem motivation, and related work were conducted manually by the authors, and all citations were selected and validated without the use of LLMs.

\noindent \textbf{Transparency.} LLMs were not used to generate research ideas, design methodology or conduct experiments. No scientific claims in this work rely on unverified LLM output. Because LLMs were used only for language refinement and not as part of the research methodology or experimentation pipeline, they do not affect reproducibility, experimental validity, or conclusions.

\noindent \textbf{Responsibility.}
No data collection or model training in this work required the use of proprietary or ethically sensitive datasets. The research does not rely on private user data, nor does it involve scraping or collecting text for the purpose of LLM fine-tuning. The experimental design emphasized efficiency by using small models, limited training epochs, and minimal GPU hours to reduce environmental impact.

In summary, LLMs were used strictly as writing assistants. All core research content of this paper is original, authored and verified by the research team, and adheres to the conference requirements on originality, transparency, and responsible use of machine learning tools.

\section*{Ethics Considerations}
This work studies model extraction attacks against graph foundation models with the goal of understanding their security properties and informing the design of safer deployment practices. Our experiments operate exclusively on publicly available datasets and pretrained models, and no private, user-specific, or sensitive data are used. All attacks analyzed in this paper are conducted in a controlled research setting for the purpose of evaluating risks and developing defenses, not for enabling misuse.

We emphasize that the methods presented here highlight potential vulnerabilities so that practitioners can better protect proprietary models, minimize unintended leakage of model capabilities, and design safer APIs. No attempt was made to access or attack any commercial, confidential, or restricted systems. Beyond these considerations, we identify no additional ethics concerns.

\end{document}